\documentclass[final,3p,times,twocolumn,authoryear]{elsarticle}
\usepackage{amssymb}
\usepackage{algorithm}
\usepackage{algorithmicx,algpseudocode}
\usepackage{algpseudocode}
\usepackage{array}
\usepackage{pdflscape}
\usepackage{longtable}
\usepackage{multirow}
\usepackage{siunitx}
\usepackage{booktabs}
\usepackage{makecell}
\usepackage{xltabular}
\usepackage{longtable}
\usepackage{afterpage}  % load the afterpage package
\usepackage{pbox}
\usepackage{tabularray}
\usepackage{placeins}
\usepackage{enumitem}
\usepackage{hyperref}
\usepackage{titlesec}
\usepackage{mathtools}

\usepackage{chngcntr} % for easy manipulation of counters
\titleformat{\paragraph}[hang]{\normalfont}{\theparagraph.}{.5em}{\emph}
\counterwithin{paragraph}{section}
\renewcommand\theparagraph{\Alph{paragraph}}
\usepackage{lineno}
\journal{Computers and Electronics in Agriculture}

\begin{document}
	
	\begin{frontmatter}

		\title{A comprehensive review of 3D convolutional neural network-based classification techniques of diseased and defective crops using non-UAV-based hyperspectral images}
	
		\author[1]{Nooshin Noshiri\corref{cor1}}
		\ead{noshiri-n@webmail.uwinnipeg.ca}
		\cortext[cor1]{Corresponding author}
		
		\author[1]{Michael A. Beck}
		\ead{m.beck@uwinnipeg.ca}
%		\cortext[cor2]{Co-author}
		
		\author[2]{Christopher P. Bidinosti}
		\ead{c.bidinosti@uwinnipeg.ca}
		\author[1]{Christopher J. Henry}
		\ead{ch.henry@uwinnipeg.ca}
		
		\affiliation[1]{organization={Applied Computer Science Department, The University of Winnipeg},%Department and Organization
			addressline={515 Portage Ave}, 
			city={Winnipeg},
			postcode={R3B 2E9}, 
			state={Manitoba},
			country={Canada}}
		\affiliation[2]{organization={Physics Department, The University of Winnipeg},%Department and Organization
			addressline={515 Portage Ave}, 
			city={Winnipeg},
			postcode={R3B 2E9}, 
			state={Manitoba},
			country={Canada}}
		
		\begin{abstract}
			%% Text of abstract
			Hyperspectral imaging (HSI) is a non-destructive and contactless technology that provides valuable information about the
 structure and composition
 of an object. It has the ability to capture detailed information about the chemical and physical properties of agricultural crops. Due to its wide spectral range, compared with multispectral- or RGB-based imaging methods, HSI can be a more effective tool for  
 monitoring
 crop health and productivity. With the advent of this imaging tool in agrotechnology, researchers can more accurately address issues related to the detection of diseased and defective crops in the agriculture industry. This allows to implement the most suitable and accurate farming solutions, such as irrigation and fertilization, before crops enter a damaged and difficult-to-recover phase of growth in the field. 
While HSI provides valuable insights into the object under investigation, the limited number of HSI datasets for crop evaluation presently poses a bottleneck. Dealing with the curse of dimensionality presents another challenge due to the abundance of spectral and spatial information in each hyperspectral cube.
State-of-the-art methods based on 1D and 2D convolutional neural networks (CNNs) struggle to efficiently extract spectral and spatial information. On the other hand, 3D-CNN-based models have shown significant promise in achieving better classification and detection results by leveraging spectral and spatial features simultaneously.
			Despite the apparent benefits of 3D-CNN-based models, their usage for classification purposes in this area of research has remained limited. This paper seeks to address this gap by reviewing 3D-CNN-based architectures and the typical deep learning pipeline, including preprocessing and visualization of results, for the classification of hyperspectral images of diseased and defective crops. Furthermore, we discuss open research areas and challenges when utilizing 3D-CNNs with HSI data.
			
		\end{abstract}

		%Research highlights
%		\begin{highlights}
%	
%		\end{highlights}
		
		
	\end{frontmatter}

	%% main text
	\section{Introduction}\label{intro}
	Plant diseases pose significant threats to global food production, with potential yield losses of up to 30\% and substantial economic impact \citep{rizzo2021plant}.
	This can have a devastating impact on farmers and communities, particularly in low-income countries where access to food is already challenging. 
	Precision agriculture and hyperspectral imaging (HSI) offer promising solutions for preventing crop damage and losses, ultimately contributing to efforts to promote sustainability and reduce the impact of diseases on food production.

HSI, also referred to as imaging spectrometry, combines two distinct technologies, imaging and spectroscopy, to provide both spatial and spectral information, simultaneously. 	
	Spectral information can provide rich information about biochemical and biophysical attributes of the agricultural crops. This is due to the higher spectral resolution of hyperspectral sensors compared to multispectral and RGB ones.
As a result, this feature
can	
lead to better discrimination of objects of similar colors, higher accuracy in complex classifications,
and the
ability to predict chemical composition and provide information about the interior of an object \citep{sun2010hyperspectral}. 
	
	However, the interpretation of spectral data can be complex, especially when analyzing and comparing multiple samples over extended periods. One approach to simplify spectral analysis is the usage of spectral indices. Spectral indices are mathematical expressions that combine several spectral bands into a single value, providing an easier representation of the data. For example, the Normalized Difference Vegetation Index (NDVI) computes the ratio between Near Infrared (NIR) and Red (R) bands of hyperspectral channels as follows: 
	\begin{equation}\label{equ0}
		\text{NDVI} = \frac{\text{NIR} - \text{R}}{\text{NIR} + \text{R}}
	\end{equation}
	With the help of spectral indices, one can effectively identify trends and changes in the data without requiring an in-depth comprehension of the underlying scientific principles governing spectral data, thus enabling a simpler data analysis, enhancement of features, standardization, comparability, and calibration of data. 
	
In the agricultural industry,
 two common spectral indices are the already mentioned NDVI and the Green Chlorophyll Index (GCI). The former is used to monitor vegetation growth and health and the latter quantifies the amount of chlorophyll in plants. Further indices have been defined, usually in the context of remote sensing, to support research for example on agriculture, soil, vegetation, water and forestry. A comprehensive database of spectral indices that is searchable by application area and hyperspectral sensor is provided in \cite{henrich2009development}.
	
\begin{figure}[h!]
		\centering
		\includegraphics[width=0.99\linewidth]{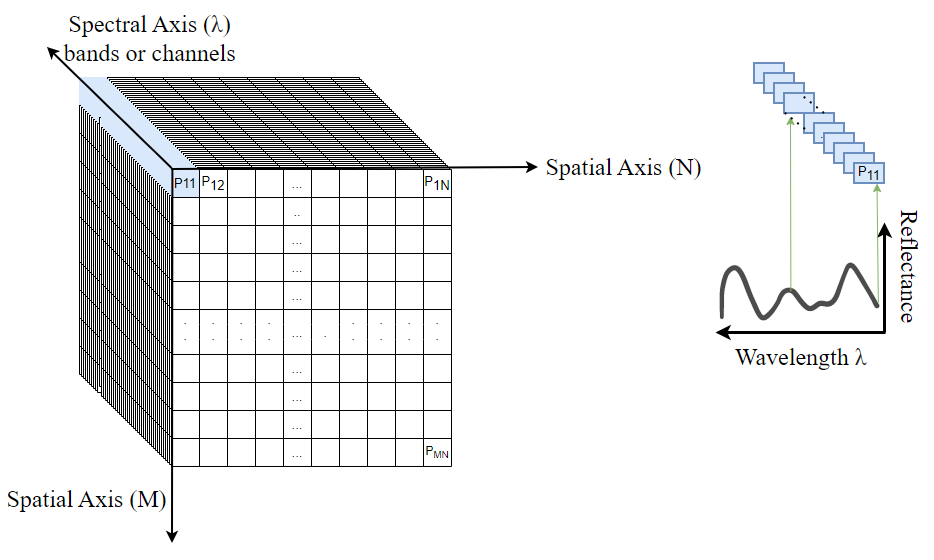} 
		\caption{The hyperspectral cube (adapted from \cite{tarabalka2010multiple} with modification). 
			It is a three-dimensional array where each pixel represents a spectrum containing a range of wavelengths. This spectrum can act as a fingerprint and provides information about biophysical and biochemical characteristics of the imaged object. 
		}
		\label{fig:cube}
	\end{figure}

	From a data-perspective, 
	 a hyperspectral image is a stack of images, known as a
	hyperspectral cube or a data cube. Each image of this cube represents the response of the imager to one of the distinct hyperspectral channels \citep{benediktsson2015spectral}. This is illustrated in Fig.~\protect\ref{fig:cube}. It shows a 3D data cube P with dimensions \(M\times N \times \lambda \), where \(M\) and \(N\) represent the axes of spatial information and $\lambda$ represents the spectral dimension \citep{tarabalka2010multiple}. In the hyperspectral cube, each pixel, given by its spatial coordinates, is a vector of length $\lambda$ that indicates the reflected radiation of a specific part of the object. 
	
	The high-dimensionality of this data cube poses a challenge to traditional machine learning approaches, resulting in reduced accuracy due to their inability to extract complex features. Moreover, the performance of these approaches heavily depends on manual feature engineering. Convolutional Neural Networks (CNN) have been proven to achieve high classification accuracies in image classification tasks and to work well with the high-dimensionality of HSI data. 
	
In this paper, we present a comprehensive review of 3D-CNN-based models utilized in the classification of non-UAV-based hyperspectral images of diseased and defective crops. This review is intended to assist computer vision experts and agriculture-domain researchers seeking to address HSI classification tasks for crops under stress.

	The paper is organized as follows. In the following Section, we outline the investigation protocol used in this review. In Section~\ref{sec:CNN}, we briefly describe the structure of CNNs and their most important concepts.
	Following the typical data pipeline associated with CNNs, 
	 Sections \ref{prep} to \ref{visualization} review data preprocessing, band and feature selection, network architecture design, and data visualization. This  
	provides a convenient overview of its individual steps for plant classification problems using 3D-CNNs.
	Finally, Section \ref{discussion} highlights the research gaps and limitations associated with the application of 3D-CNNs for HSI data classification.

	\section{Search methodology}\label{material}
	
	A systematic search was conducted by accessing scholarly publications through google scholar search engine. To optimize the search results, specific keywords were employed in the advanced search section, resulting in a refined list of articles. The selected search terms were different combinations of ``hyperspectral'', ``disease'', ``detection'', ``identification'', ``diagnosis'', ``plant'', ``crop'', ``stress'', ``3D CNN'', ``3 dimensional CNN'', ``three dimensional CNN'', utilizing the Boolean operators, AND and OR. Around 2,000 records were investigated, however, to ensure the relevance of the articles, the abstracts of the retrieved papers were thoroughly assessed to confirm coherence with the title of this research. Moreover, screening criteria were implemented, including the removal of non-English papers, to ensure an accessible and high-quality selection of articles. The results of this comprehensive study are provided in Table \ref{table1} and Table \ref{table2} (see Section \ref{net}) and contain papers from 2015 up to February of 2023.

%NN	
This review examines various applications of 3D-CNN-based models in detecting and classifying diseases in agricultural crops, including charcoal rot in soybeans \citep{nagasubramanian2018explaining}, mold in peanuts and strawberries \citep{liu2020using, jung2022hyperspectral}, bacterial leaf blight (BLB) in rice \citep{cao2022detecting}, grapevine vein-clearing virus (GVCV) in grapevines \citep{nguyen2021early}, and potato late blight (PLB) in potatoes \citep{qi2023field}. 
	Moreover, the review explores the use of 3D-CNN-based architectures for identifying specific defects in crops, such as decay in blueberries \citep{qiao2020detection}, bruise and brown spots in fruits \citep{pourdarbani2023examination,jia2023net}, heat stress in rice \citep{gao2021hyperseed}, as well as black, fermented, shell, and broken coffee defects in beans \citep{chen2022real}. By harnessing the power of 3D-CNN-based models, we can effectively address challenges of classifying diseased and defective using hyperspectral images. This can result in the preservation of product quality, prevention of yield losses, and ensuring food safety standards.

	\section{CNN structure and concepts} 
	\label{sec:CNN}

\begin{figure*}[h!]
		\centering
		\includegraphics[width = \textwidth, height = 4cm, keepaspectratio]{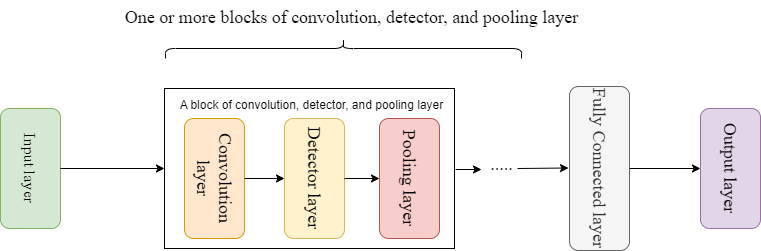} 

		\caption{A basic conceptual CNN architecture. 
			A CNN consists of multiple layers, including convolution, detector, and pooling layers, where each layer serves as the input for the subsequent layer, enabling the extraction of low-dimensional features, learning of non-linear representations, and dimensionality reduction in the network.	
		}
		\label{fig:cnn}
	\end{figure*}

\begin{figure}[h!]
		\centering
		\includegraphics[width = \textwidth, height = 12cm, keepaspectratio]{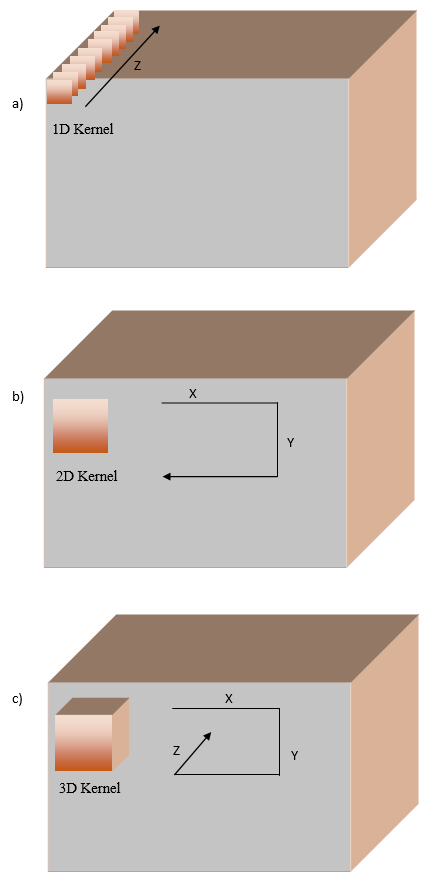} 

		\caption{Movement direction of convolution process using 1D, 2D, and 3D kernels in hypercube. A schematic overview of movement directions for (a) 1D, (b) 2D, and (c) 3D convolutions in CNNs over hypercube is shown. The X and Y directions indicate the movement of the kernel across the spatial dimensions, and Z direction shows the movement across the spectral dimension.
		}
		\label{fig:conv}
	\end{figure}

	A CNN is comprised of a series of
	layers 
	each consisting of several neurons. 
	As shown in Fig.~\protect\ref{fig:cnn}, each layer is the input for the next layer in the network. Key building blocks of a CNN are convolution layers, detector layers, and pooling layers. A convolution layer uses convolutional kernels to extract low-dimensional features from the input data while preserving the spatial relationship between the input data pixels. Fig.~\ref{fig:conv} depicts the movement directions of 1D (spectral), 2D (spatial), and 3D (spatial-spectral) kernels  within the hypercube. A detector layer applies a non-linear function like Rectified Linear Unit (ReLU) to learn non-linear representations. Pooling layers make features invariant by reducing the dimensionality of data. 
	
	Classification methods based on 1D-CNN (spectral feature-based) and 2D-CNN (spatial feature-based) cannot efficiently classify hyperspectral data as neither utilizes spatial and spectral features together. However, a 3D-CNN can extract spatial-spectral features from the volumetric data. This is due to its ability to incorporate the spectral dimension in addition to the spatial dimensions, which enables it to model and learn more complex spatiotemporal representations.

	In the following, a description of previous review papers that deal with
	the topics of CNNs, hyperspectral data, and applications in agriculture and plant research is given. The work of \cite{signoroni2019deep} is aimed at domain professionals seeking comprehensive insights into the integration of hyperspectral acquisition techniques and deep learning architectures for specific tasks across diverse application domains. This resource caters to machine learning and computer vision experts, offering a nuanced understanding of how deep learning technologies are tailored to effectively process and analyze hyperspectral data, keeping them up-to-date with the latest advancements. In \cite{jiang2020convolutional} an examination of how different CNN architectures have been employed in the assessment of plant stress, plant development, and postharvest quality. This review categorizes the studies according to the technical advancements achieved in terms of imaging classification, object detection, and image segmentation. As a result, it highlights cutting-edge solutions for specific phenotyping applications, offering valuable insights into the current state of the field. An overview of state-of-the-art CNN models and visualization techniques for disease diagnosis in plants is given in \cite{joseph2022intelligent}. The review given in \cite{gill2022comprehensive} delves into the plant stress phenotyping, specifically examining the utilization of machine learning and deep learning methodologies. The study encompasses a wide range of high throughput phenotyping platforms, exploring the integration of data from diverse sources. 
	However, it does not discuss 3D-CNN architectures. Finally, a review paper that combines all three topics was provided in \cite{wang2021review}. 
	In that work, the authors provide an overview of the application of hyperspectral imaging in agriculture, encompassing areas such as ripeness and component prediction, classification themes, and plant disease detection. Additionally, the study examines recent advancements in hyperspectral image analysis specifically in the context of deep learning models. The review not only highlights the achievements in this field but also outlines the existing challenges associated with deep learning-based hyperspectral image analysis. Moreover, the study presents future prospects and potential directions for further research in this domain.
	
	In this review, we update and complement the paper of \cite{wang2021review} with a particular focus on 3D-CNNs and the entirety of the process of creating a high-performant model, including preprocessing of data, band selection, exploration of model architectures, and data visualization.
	
	\begin{figure}[h!]
		\centering
		\includegraphics[width = \textwidth, height = 15cm, keepaspectratio]{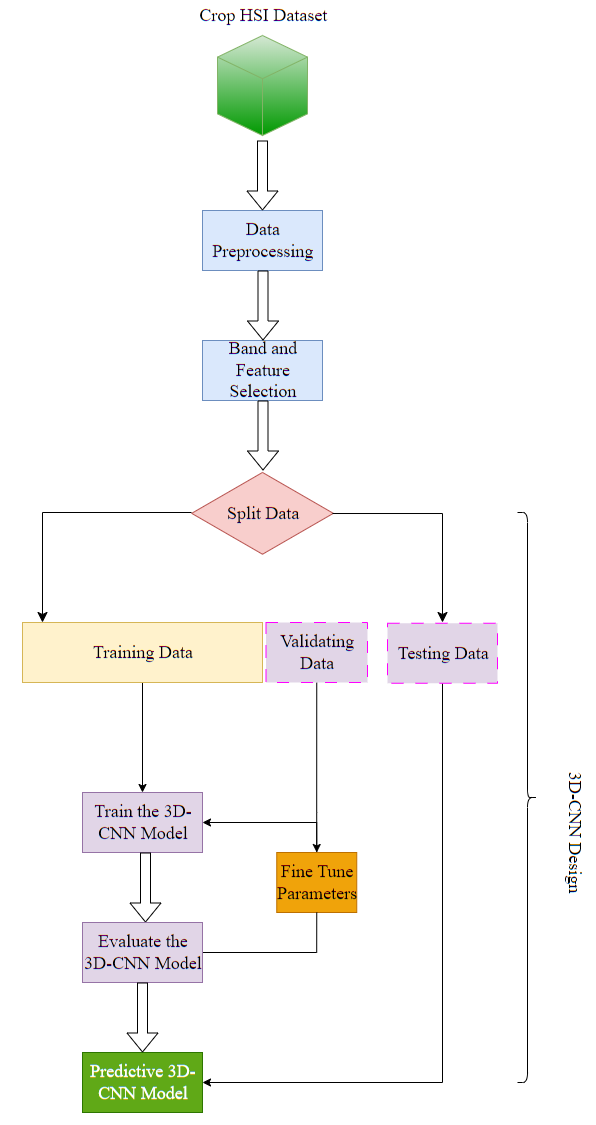} 
		\caption{Typical deep learning pipeline for HSI data classification. First, the discriminative features and bands are extracted from preprocessed HSI dataset of crop. Then, the dataset is split to training, validation, and testing sets. The training dataset is used to train the 3D-CNN model, the validation dataset is employed to assess the model's performance and fine tune its parameters, and the testing dataset serves to evaluate the final performance and generalization ability of the trained 3D-CNN model on unseen data.
		}
		\label{fig:hsi_cnn}
	\end{figure}

	A deep learning pipeline for HSI classification typically consists of several stages, including data preprocessing, band and feature selection, model design, model training, testing, and evaluation (see Fig.~\ref{fig:hsi_cnn}). Data preprocessing involves enhancing the quality of raw HSI data, for example, through noise removal, radiometric calibration, and dimension reduction. 
	Feature extraction transforms the raw data into a new space of features, which is expected to be more discriminative for the classification task. Band selection defines a subset of the original spectral bands that is the most relevant for the classification task. Model design is the configuration of hyperparameters of the CNN model, for example, the number and sequence of convolutional layers and dense layers, the activation functions to be used, and the usage of dropout or skip connections. During the training step, the model is optimized by iteratively adjusting its internal parameters (weights and biases) to minimize the error between its predicted output and the actual output on the input data. Finally, the model is evaluated to learn patterns and relationships in the data and to make accurate predictions on new unseen data.

	\section{Data preprocessing}\label{prep}
	
	Data preprocessing is a critical step in hyperspectral data analysis, aimed at optimizing the quality and quantity of the data. This step enhances the suitability of the data for downstream tasks such as classification and feature extraction. Preprocessing techniques include patch extraction, radiometric correction and calibration, smoothing, dimension reduction, and background removal. Data augmentation also falls under preprocessing and has the goal of increasing the volume of training data.
	
	\subsection{Patch extraction}\label{patch}
	
	Patch extraction is a technique that involves dividing an image into smaller images or patches. In the context of HSI, patch extraction has significant advantages for efficient and targeted analysis of specific regions of interest within an image. By extracting image patches that contain pixels with similar properties, researchers can focus their analysis on areas of the image that are most relevant to diseased or defective areas. 
	
	To give a concrete example, \cite{nagasubramanian2018explaining} utilized patch extraction to analyze hyperspectral images of soybean crops. In their study, they extracted spatial patches of size 64$\times$64$\times$240 from an original image of size 500$\times$1600$\times$240, where the first two dimensions define the spatial resolution of the image and 240 denotes the number of spectral bands. By analyzing the properties of pixels within these patches, they were able to extract features that were more representative of the target disease which ultimately improved the accuracy and efficiency of their analysis. Furthermore, patch extraction helps to expand the number of images when there is a lack of data. It also reduces computational time, as processing hyperspectral images of large sizes can be computationally demanding \citep{qi2023field,jia2023net}.
	
	\subsection{Data augmentation} \label{aug}
	
	Imbalanced data is a common problem in many machine learning applications and HSI is no exception. It refers to a situation where the number of samples in each class or category of the data is not evenly distributed. This can lead to a bias towards the overrepresented classes in the analysis results, which is particularly problematic if the minority class is of interest.
	
	To address imbalanced HSI data, one common approach is resampling \citep{nguyen2021early}, which involves either oversampling the minority class or undersampling the majority class to balance the class distribution. Resampling can be done randomly or using more advanced techniques such as Synthetic Minority Oversampling Technique (SMOTE) \citep{chawla2002smote} or Adaptive Synthetic (ADASYN) sampling \citep{he2008adasyn}.
	Other widespread data augmentation approaches
	include
	transformation techniques
	such as
	mirroring \citep{liu2020using},
	rotation \citep{liu2020using, chen2022real}, horizontal and vertical flipping \citep{chen2022real, pourdarbani2023examination}, and
	color jittering \citep{pourdarbani2023examination}.
	Along with above-mentioned methods, patch extraction (Section \ref{patch}) can also be used to address imbalance HSI dataset. 
	
	\subsection{Radiometric calibration and correction}
	
	Radiometric calibration is an essential step in the accurate operation of hyperspectral cameras. It aims to establish a quantitative relationship between the response of the camera sensor (the radiation sensor) and the actual reflectance (radiation level) of an object in a given environment. The calibration process involves assigning a
	``true'' 
	value for either radiation intensity or reflectance to the digital numbers given by the camera that represent recorded outputs for each pixel and spectral channel. To that end, calibrated reflectance standards are utilized, which usually consist of a matte Lambertian reflecting surface to ensure that reflected light is uniform in all directions \citep{shaikh2021calibration}. 
	
	The calibration process is then performed as follows:
	After selecting a radiation source that emits a known type and amount of radiation, the detector is placed in close proximity to the source to measure the radiation. Next, using calibration factors provided by the manufacturer or based on known mathematical formulas (see, for example, \citep{qi2023field}) for the particular detector, the expected response of the detector is determined. 
	The measured response of the detector is then aligned with the expected response based on the known radiation level using calibration factors.
	This calibration process should be repeated periodically with different radiation sources to ensure the continued accuracy and reliability of the detector's measurements.
	
\begin{figure*}[h!] 
		\centering
		\includegraphics[width = \textwidth, height = 6cm, keepaspectratio]{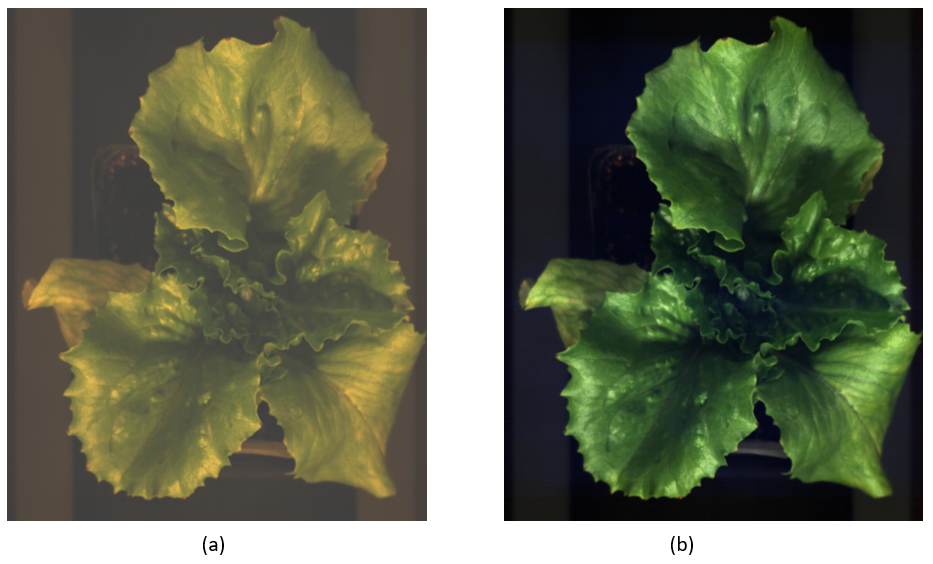}
		\caption{Black and white calibration of HSI data. (a) The hyperspectral image of lettuce rendered before black and white image correction; (b) The image shows the rendered hyperspectral image after black and white image correction using the Equation \ref{equp}.    
		}
		\label{fig:stripe}
	\end{figure*}

	Even with calibration performed, hyperspectral cameras are susceptible to radiometric errors that can arise from a variety of sources, including sensor drift, electronic interference, light source, and data transmission and recording issues. 
For example, one of the common radiometric errors which relates to the camera's sensor is stripping. Each sensor consists of multiple individual detectors that sometimes do not function properly due to being out of calibration. Consider a push broom (along-track) camera where its sensor has multiple detectors aligned in a row. When one of them is calibrated slightly different from the adjacent detector the striping effect can occur. In this case, lines predominantly consisting of varying shades of dark and bright pixels formed.
	Radiometric correction can register and rectify
	incorrect
	pixel brightness.
	To achieve this, a series of procedures are employed including noise correction, de-striping, line-dropout correction \citep{duggal2013surveying}, and black and white image correction \citep{gao2021hyperseed, cao2022detecting, chen2022real,jia2023net}.
Figure~\ref{fig:stripe}(b) shows the calibrated hyperspectral image of lettuce using the following computation,
\begin{equation}\label{equp}
	\text{Calibrated HSI Data} = \frac{\text{Raw HSI Data} - \text{Dark Ref}}{\text{White Ref} - \text{Dark Ref}},
\end{equation}
where, Raw HSI Data is the image taken by the hyperspectral camera without modification, Dark Ref is the image captured by closing the camera's lens, and the White Ref is the imaged white reference with even and maximum reflectance across the spectral range.    
	
	\subsection{Smoothing} \label{preprocessing_smoothing}
	
	Smoothing, also known as filtering, is a technique in HSI to improve the quality of the data by reducing noise and artifacts and enhancing the signal-to-noise ratio of the data. This technique in HSI can be broadly classified into two categories: spatial smoothing and spectral smoothing.
	
	Spatial smoothing techniques selectively smooth out certain features in an image by applying filters that amplify or attenuate certain spatial frequencies \citep{duggal2013surveying}. 
	Common spatial smoothing techniques include Gaussian filtering, mean filtering \citep{rees2013physical}, median filtering, bilateral filtering \citep{cao2017hyperspectral}, and anisotropic diffusion filtering \citep{lennon2002nonlinear}. 
	
	Spectral smoothing techniques operate on the spectral domain of an HSI image by smoothing the intensity values of neighboring spectral bands. As in spatial smoothing the goal is to remove noise \citep{vaiphasa2006consideration}, reduce artifacts, and enhance features in the image. Common spectral smoothing techniques include moving average filtering \citep{vaiphasa2006consideration}, Savitzky-Golay (SG) filtering \citep{vidal2012pre}, Fourier filtering, wavelet filtering, and principal component analysis (PCA). For example, \cite{cao2022detecting} eliminated the random noise that was present in the spectral data of the different
	region of interests (ROIs) using SG filtering. 
	The filtering process makes it easier to identify the true signal of the sample and remove interference caused by size and structure differences between the ROIs.
	Similarly, \cite{jung2022hyperspectral} and \cite{jia2023net} used a SG filter 
	to smooth out the spectral data and reduce the effect of noise.
	
	As \cite{vaiphasa2006consideration} points out, care must be taken when applying smoothing techniques. The subjective selection of smoothing filters in hyperspectral remote sensing studies can negatively impact the statistical properties of the spectral data, which can, in turn, affect subsequent analyses. To preserve the statistical properties of the HSI data, the selection of smoothing filters should be done through a comparative t-test method that identifies the filter with the least statistical disturbances. By using this approach, it is possible to mitigate the negative effects of smoothing filters and ensure the reliability of subsequent analyses based on statistical class models.
	
	\subsection{Dimension reduction}
	
	The large number of spectral bands within hyperspectral data often makes it challenging to process and analyze HSI data. Dimensionality reduction techniques retain relevant information, while allowing the model to work on smaller hypercubes downstream. These techniques are split into linear and non-linear ones. 
	
	\subsubsection{Linear techniques}\label{dimension_reduction:linear}
	
	Linear dimension reduction in HSI refers to a set of statistical and machine learning techniques that aim to find a lower-dimensional linear
	manifold in the high-dimensional space that captures the essential spectral information of the data \citep{khodr2011dimensionality}. By embedding the data into this lower-dimensional linear space, the dimensionality of the data is reduced while preserving as much of the spectral information as possible.
	PCA and Random Forest (RF) are two
	examples
	of this technique,
	and demonstrations of both can be seen in 
	\cite{cao2022detecting}. 
	
	The process of PCA involves first zero-centering the input spectral matrix (i.e., the matrix of the spectral bands) and computing its covariance matrix. Next, the eigenvectors and eigenvalues of the covariance matrix are calculated, and the eigenvalues are sorted in descending order. The top $k$ eigenvalues are then selected and the corresponding eigenvectors space computed. The $k$-dimensional data is obtained by projecting the original spectral matrix into the new space using the selected eigenvectors.
	
	On the other hand, an RF evaluates the importance of each wavelength by randomly replacing it and measuring its effect on the accuracy of the thus trained CNN model. 
	To accomplish this, the RF builds many decision trees, and each tree is trained on a different subset of the data. The algorithm calculates the importance score for each wavelength by measuring the change in the prediction error rate before and after randomly replacing the wavelength in the out-of-bag data. The wavelength with the highest importance score is selected, and this process is repeated until the desired number of wavelengths is achieved. The dimensionality of the hyperspectral data is thus reduced, while retaining the most important information for accurate predictions.
	
	Further linear algorithms for dimension reduction of HSI data are described in \cite{firat2022classification} and encompass Independent Component Analysis (ICA) and PCA-based algorithms like Incremental PCA (IPCA), Sparse PCA (SPCA), and Randomized PCA (RPCA).
	
	\subsubsection{Non-linear techniques} 
	
	Non-linear techniques can handle data with complex and non-linear structures. \cite{Yang2009} classified these techniques into 
	Kernel-based and manifold learning \citep{Yang2009}. 
	Kernel-based techniques, like Kernel PCA (KPCA), use non-linear mappings to transform data into higher-dimensional feature spaces first on which linear techniques can be applied. Manifold learning algorithms, on the other hand, aim to directly discover the intrinsic non-linear structure of data. See \citep{khodr2011dimensionality} for well-known manifold learning techniques, including Isometric Feature Mapping (Isomap), Locally Linear Embedding (LLE), Local Tangent Space Alignment (LTSA), Diffusion Maps, Sammon's Mapping (SM), and Locality Preserving Projections (LPP).
	
	\subsection{Background removal}
	
	The data captured in hyperspectral images contains both foreground, called the
	ROI, and background objects. In scenarios where the target object does not cover the entire scanning area the signals from background objects can interfere with the data analysis, i.e., the background can contain noise that needs to be filtered out \citep{vidal2012pre}. This is especially true when dealing with images that exhibit color gradients. 
	By masking the background from the data, researchers can focus on the spectral signature of the ROI. This leads to improved target detection in classification task. It also reduces the computational complexity of subsequent processing steps including the training 3D-CNN models \citep{qi2023field}.
	
	There are various traditional techniques to extract he ROI from the hyperspectral image. These techniques can be classified into several categories based on their underlying principles
	and are discussed separately below.
	
	\subsubsection{Spectral similarity-based methods} 
	
	These methods work based on the similarity between the spectra of the pixels within an image. Examples of such methods include Spectral Angle Mapper (SAM) \citep{kumar2015comparison} and Spectral Information Divergence (SID) \citep{qin2009detection}.
	SAM computes the spectral angle 
	\begin{equation}\label{equ7}
		\alpha = \cos^{-1} \left(\frac{\sum_{i=1}^{\lambda} R_{i}\cdot T_{i}}{(\sum_{i=1}^{\lambda} R_{i}^{2})^{\frac{1}{2}}\cdot(\sum_{i=1}^{\lambda} T_{i}^{2})^{\frac{1}{2}}}\right)
	\end{equation}
	between the target reference spectrum $R$ and each pixel spectrum $T$ for all spectral bands $\lambda$ in the hyperspectral image. This results in a similarity measure that is insensitive to illumination variations \citep{avbelj2012spectral}. On the other hand, SID works based on the concept of information theory (entropy). It compares the spectral information content of each pixel to a reference spectral information content. In general, SAM is better suited for well-defined spectral variations and low background noise, while SID is more robust to complex background noise and illumination variations.

	\subsubsection{Statistical-based methods} 
	
	These methods leverage statistical techniques to identify ROIs that share similar spectral properties and produce a set of uncorrelated components capturing different aspects of spectral variability within the image. For example, Minimum Noise Fraction (MNF) \citep{luo2016minimum} is specifically designed to reduce the impact of noise in the data by separating the noise and signal components of the HSI data. This makes it particularly useful for operating on noisy HSI data but at the expense of more computational time. Another technique, ICA (see Section \ref{dimension_reduction:linear}), aims to separate the mixed signals into their independent components, providing a more flexible approach to identify subtle spectral differences between ROIs. 
	Finally, PCA \citep{chen2022real} decomposes the original HSI data into orthogonal components that represent the directions of maximum variance in the data. In general, PCA is computationally more efficient than ICA and MNF, as it involves a simpler mathematical transformation that uses standard matrix operations. 
	
	\subsubsection{Spatial-based methods} 
	
	These methods exploit the spatial correlation present in the image to differentiate between pixels within and outside the ROI. Typically, these methods apply morphological operations or spatial filtering techniques to extract features such as edges or texture, which are then used to segment the image into ROIs. The Morphological Attribute Profile (MAP) \citep{dalla2010morphological} and the Spatial-Spectral Endmember Extraction (SSEE) algorithm \citep{plaza2011recent} are examples of spatial-based methods that use mathematical morphology and spatial filtering, respectively. 
	
	SSEE first identifies the endmembers, or pure spectral signatures, present in the HSI data, and then uses a spatial clustering algorithm to group adjacent pixels with similar spectral properties into ROIs. On the other hand, MAP applies a series of morphological opening and closing operations to the image to identify connected regions of pixels with similar morphological attributes, such as size and shape. These connected regions can then be used as ROIs. These methods are computationally efficient and can be useful in scenarios where spectral information alone is not sufficient for accurate ROI extraction, such as in cases of low spectral contrast or high noise levels.
	
	\subsubsection{Hybrid methods} 
	
	These methods combine techniques to improve the accuracy and efficiency of the ROI extraction process. One approach can be combining statistical-based techniques such as PCA, ICA, or MNF with spatial-based techniques such as MAP or SSEE.
	For example, by combining MAP and PCA we can exploit both the spatial and spectral information \citep{sun2021patch} in the HSI data for ROI extraction. These hybrid methods can be effective in cases where neither spatial nor spectral methods alone are sufficient for accurate ROI extraction.
	
	\subsubsection{Machine learning-based methods}
	
	Machine learning algorithms can also be used to extract the ROI. This process involves training a model using labeled data to identify and extract regions with specific spectral characteristics. Once trained, the model can be used to predict the presence of those characteristics in unlabeled data and extract ROIs. This procedure is computationally efficient and allows for the extraction of subtle and complex patterns that may not be easily identifiable through traditional methods. Some instances of such techniques are Support Vector Machines (SVM) \citep{bojeri2022automatic}, RF \citep{boston2022comparing}, and CNN \citep{li2022semantic,wan2023yolo}. 
	
	\subsubsection{Software-assisted manual annotation}
	
	Manual definition of the ROI can be assisted by software specifically built to handle HSI data. Amongst these are for example ENVI \citep{HG} and Spectronon \citep{RI} as commercial products, as well as the MATLAB Hyperspectral toolbox. Examples for open-source and free software are SeaDAS \citep{NASA}, the Orfeo ToolBox \citep{CNES}, and RSGISLib \citep{Bunting_etal_2014}. These packages provide HSI data analysis for a wide range of tasks discussed before.
	An example of software-assisted annotation is 
	the work of \cite{jin2018classifying} in which the authors generated an ROI using ENVI by manually selecting the tissues or areas of interest in the false color images of HSI data.
	
	Researchers can utilize such software in addition to the above discussed algorithms. For example, \cite{gao2021hyperseed} developed an open-source software which is specifically designed for analyzing HSI data of seeds. It is able to remove the background of the HSI data and produces a binary mask using a user-defined minimum and maximum intensity threshold along with a component-searching algorithm. The results are an accurate segmentation of the seeds even when they overlap.

	\section{Band and feature selection }\label{bands}
	
	The selection of spectral bands in HSI is another crucial preprocessing step for classification tasks. It entails the identification of a subset of the most discriminative spectral bands from all available bands. This process is instrumental in reducing data dimensionality by eliminating redundant bands, thereby significantly reducing the computational overheads of downsteam tasks.
	Furthermore, the careful selection of spectral bands can also reduce the effects of noise in the data. In general, based on the survey of \cite{sun2019hyperspectral}, band selection mechanisms can be categorized into six groups as follows (another classification of methods is presented in \cite{sawant2020survey}).
	
	\subsection{Ranking-based selection}
	
	Ranking-based band selection methods evaluate the significance of each spectral band based on a predetermined criterion and choose the most important bands in a sorted order. These methods can be categorized into two types: supervised and unsupervised. In supervised ranking-based methods, labeled training samples are utilized to determine the importance of each spectral band, while in unsupervised ranking-based methods, statistical properties of the data are used for the same purpose.
	
	One example of such method is spectral differentiation. \cite{qi2023field} employed first and second order differentiation to decrease the computational complexity of HSI data containing 204 bands. The first derivative can effectively pinpoint areas in the spectrum where the rate of change is highest. This indicates the presence of sharp spectral features such as absorption or emission lines. Moreover, it allows for the selection of bands that capture these features and thus, provide critical information for classification or detection tasks.
	The second derivative is useful for identifying regions of the spectrum where the rate of change of the first derivative is highest, signifying the presence of spectral curvature. The selection of bands that capture the shape of the spectral signature, based on this information, can enhance the differentiating power in classification or detection tasks. Likewise, \cite{jung2022hyperspectral} achieved an increase in accuracy using spectral differentiation and expansion of the input in the vertical direction of the raw data. This technique was applied in addition to SG smoothing (see Section \ref{preprocessing_smoothing}) to further improve data quality.
	
	\subsection{Searching-based selection}
	Searching-based band selection methods involve the creation of a criterion function such as Euclidean distance and Bhattacharyya distance \citep{ifarraguerri2004visual} to evaluate the performance of each spectral band based on a specified optimization objective. The first step involves creating an initial subset of bands, followed by an assessment of the criterion function for the subset. The next step is applying a searching strategy to identify the best subset of bands that maximizes the criterion function, and evaluating the selected subset based on data classification performance. This iterative process continues until the desired level of performance is reached. Searching-based methods largely depend on the quality of the criterion function and the optimization strategy employed. Incremental searching \citep{wang2007novel,dos2015efficient}, updated searching \citep{ghamisi2014novel, shi2016hyperspectral}, and eliminating searching \citep{sun2014efficient} are among the commonly utilized strategies.
	
	\subsection{Clustering-based selection}
	Clustering-based methods for hyperspectral band selection group bands into clusters and select representative bands from each cluster to create a final subset. These algorithms can be unsupervised \citep{imbiriba2015band,yang2017discriminative}, supervised \citep{mojaradib2008novelband} or semisupervised \citep{su2011semisupervised,su2012adaptive}. The selection of representative bands is typically performed using information measurements, such as mutual information or Kullback–Leibler divergence. Commonly used clustering techniques are $K$-means, affinity propagation, and graph clustering. $K$-means selects the best cluster centers that minimize the sum of distances to a set of putative center candidates. Affinity propagation selects exemplars by considering the correlation or similarity among bands and the discriminative capability of each band. 
	
	\subsection{Sparsity-based selection}
	Sparsity-based techniques for band selection rely on sparse representation or regression to identify representative bands.
	The most common of these methods are discussed separately below.
	
	\subsubsection{Sparse Nonnegative Matrix Factorization-based methods} 
	
	These methods \citep{li2011clustering} break down the hypercube into a set of building blocks, which are both nonnegative and sparsely encoded. This promotes a feature extraction process that combines these building blocks to create a parts-based representation of the original data. The goal of this method is to identify the most informative bands of the HSI data matrix by optimizing an objective function that includes sparsity constraints.
	
	\subsubsection{Sparse representation-based methods} 
	Sparse representation-based methods \citep{zhai2016squaring,sun2017fast} use pre-defined or learned dictionaries to select informative bands of the HSI data matrix based on their sparse coefficients. These methods rank the bands according to the frequency of their occurrence in the sparse coefficient histograms. In some cases, sparse representation-based methods can also be designed to solve multiple tasks simultaneously, and an immune clonal strategy can be used to search for the best combinations of informative bands.
	
	\subsubsection{Sparse regression-based methods} 
	Sparse regression-based techniques \citep{sun2014new,damodaran2017sparse} transform the band selection problem into a regression problem and estimate the most representative bands by solving a sparse regression problem. These methods can also include sparsity constraints to encourage the selection of only the most informative bands for the regression model.

	\subsection{Embedding-learning based selection}
	Embedding-learning based methods aim to learn a low-dimensional representation of the spectral data, also known as an embedding, that captures the most salient features of the data.
	There are several types of embedding learning-based methods that can be used for band selection, including autoencoders \citep{tschannerl2018segmented}, Deep Neural Networks (DNN) \citep{zhan2017hyperspectral}, and CNNs \citep{sharma2016hyperspectral}. Autoencoders learn a compact representation of the input data by training a neural network to encode the input into a lower-dimensional space and then decode it back into the original space. DNN, on the other hand, consists of multiple layers of interconnected neurons. This architecture enables the network to learn complex patterns and features in a hierarchical manner, empowering it to extract high-level representations from the input data. Meanwhile, CNNs can learn spatially invariant features from image data.
	
	These techniques aim to learn a set of parameters that minimize an objective function that measures the model's performance on a particular task, such as classification or target detection. Band selection is integrated into the optimization process by constraining the learning algorithm to focus on a subset of the available bands, or by assigning weights to each band that reflect its relevance to the task at hand. The resulting model can then be utilized to predict the class label of new samples or to detect the presence of specific targets in the image.
	For example, \cite{chen2022real} employed a
	DNN based binary classification to identify the foreground and background regions of the HSI data. Subsequently, connected component labeling algorithms and edge contours were used to isolate the ROI from the image for further analysis. 
	
	Moreover, \cite{jia2023net} implemented a CNN-based band selection module that works based on a group convolution technique, which involves applying a 1$\times$1 one-dimensional convolution (equivalent to a scalar multiplication) to each band of the input hyperspectral image independently. This technique helps to overcome the problem of mutual interference between different channels. The weights of the convolutional kernel are updated in the early stage of the network training using a loss function and an auxiliary classifier (see also the next Section \ref{net}). The weights represent the importance of each band, with a higher absolute value of weight indicating greater importance of the corresponding band. 
	
	\subsection{Hybrid-scheme based selection}
	Hybrid-scheme based methods involve combining multiple band selection techniques to select the most appropriate bands. A popular combination is clustering and ranking \citep{yin2010optimal,datta2015combination}, where clustering is used to group bands and ranking is used to select the most important bands within each cluster. Other hybrid methods combine clustering with searching or combine ranking with searching to further optimize band selection. 
	
	\section{Network architecture design: Feature extraction and classification} \label{net}
	
	\begin{table*}[t]
		\caption{Non-hybrid 3D-CNN-based architectures for detection of diseased and defected hyperspectral images of crop.}
		\label{table1}
			\begin{tabular}{>{\centering\arraybackslash}m{3cm} m{9cm} >{\centering\arraybackslash}m{3cm}}		
				\hline
				\textbf{CNN Model}    
				& 
				\textbf{Information}
				&
				\textbf{Reference}\\
				
				\hline
				
				3D-CNN-based
				&
				\textbf{Dataset}: 111 hyperspectral images of soybean 
				\newline
				\textbf{Type of Disease: }Charcoal rot
				\newline
				\textbf{Imaging Device}: Pika XC hyperspectral line imaging scanner
				\newline
				\textbf{Spectral range:} 400–1000 nm
				\newline
				\textbf{GPU: }NVIDIA Tesla P40		
				&
				\cite{nagasubramanian2018explaining}

				\\
				\hline
				3D-CNN based on AlexNet
				&
				\textbf{Dataset}: 40 hyperspectral images of Grapevine groups
				\newline
				\textbf{Type of disease: }GVCV	 
				\newline
				\textbf{Imaging Device}: SPECIM IQ
				\newline
				\textbf{Spectral range: }400–1000 nm 
				\newline
				\textbf{GPU: }-		
				&
				\cite{nguyen2021early}
				
				\\
				\hline
				HyperSeed
				&
				\textbf{Dataset}: 200 rice seeds (274,641 pixels)
				\newline
				\textbf{Type of defect: }Heat stress
				\newline
				\textbf{Imaging Device}:Micro-Hyperspec Imaging Sensors, Extended VNIR version
				\newline
				\textbf{Spectral range: }600-1700 nm
				\newline
				\textbf{GPU: }-
				&
				\cite{gao2021hyperseed}
				
				\\
				\hline
				3D-CNN-based
				&
				\textbf{Dataset}: Above 200 strawberry leaves (3,110 ROIs) 
				\newline
				\textbf{Type of disease: }Gray mold 
				\newline
				\textbf{Imaging Device}: Corning microHSI
				\newline
				\textbf{Spectral range: }400-1000nm
				\newline
				\textbf{GPU: }NVIDIA RTX3090
				X (24 GB memory)
				&
				\cite{jung2022hyperspectral}

				\\
				\hline

			\end{tabular}

		\end{table*}%
		
		\begin{table*}[t!]
			\caption{Hybrid 3D-CNN-based architectures for detection of diseased and defected hyperspectral images of crop.}
			\label{table2}
				\begin{tabular}{>{\centering\arraybackslash}m{3cm} m{9cm} >{\centering\arraybackslash}m{3cm}}
					\hline
					\textbf{CNN Model}    
					& 
					\textbf{Information}
					&
					\textbf{Reference}\\
					
					\hline

					Hypernet-PRMF

					&
					\textbf{Dataset}: 16 hyperspectral images of peanut
					\newline
					\textbf{Type of disease: }Mold
					\newline	
					\textbf{Imaging Device}: SOC710E portable hyperspectral imager
					\newline
					\textbf{Spectral range:} 400–1000 nm
					\newline
					\textbf{GPU: }NVIDIA Tesla
					P100 GPU (12G)
					&
					\cite{liu2020using}
					
					\\
					\hline
					Deep ResNet 3D-CNN
					&
					\textbf{Dataset}: 16,346 hyperspectral images of blueberry
					\newline
					\textbf{Type of defect: }Distinguishing decayed and sound blueberries
					\newline
					\textbf{Imaging Device}: -
					\newline
					\textbf{Spectral range: }400-1000nm
					\newline
					\textbf{GPU: }-
					&
					\cite{qiao2020detection}

					\\
					\hline

					SDC-3DCNN
					&
					\textbf{Dataset}: Rice leaves (Number of taken samples are not determined.)
					\newline
					\textbf{Type of disease: }BLB
					\newline
					\textbf{Imaging Device}: Raptor EM285
					\newline
					\textbf{Spectral range: }378.28–1033.05 nm
					\newline
					\textbf{GPU: }NVIDIA GeForce RTX 2080Ti GPU and the AMD Ryzen 5-1600 Six-Core processor @ 3.20 GHZ CPUs
					&
					\cite{cao2022detecting}
					
					\\
					\hline
					
					2D-3D-CNN (Defect detection module of RT-CBDIA)
					&
					\textbf{Dataset}: 1026 coffee beans
					\newline
					\textbf{Type of defect: }Black, insect-damaged, and shell
					\newline
					\textbf{Imaging Device}: Imec XIMEA snapshot sensor
					\newline
					\textbf{Spectral range: }660-980 nm
					\newline
					\textbf{GPU: }GPU of GEFORCE GTX1660 Ti and a RAM of 16 GB
					&
					\cite{chen2022real}
					
					\\
					\hline

					ResNet
					&
					\textbf{Dataset}: 210 lemons
					\newline
					\textbf{Type of defect: }Bruise  
					\newline
					\textbf{Imaging Device}: It was not determined, however was provisioned by \href{https://hyperspectralimaging.ir}{Noor Imen Tajhiz Co.}. 
					\newline
					\textbf{Spectral range: }400-1100 nm
					\newline
					\textbf{GPU: }Trained on Google Colab
					&
					\cite{pourdarbani2023examination}

					\\
					\hline
					PLB-2D-3D-A
					&
					\textbf{Dataset}: 15,360 potato leaves
					\newline
					\textbf{Type of disease: }PLB
					\newline
					\textbf{Imaging Device}: Specim IQ
					\newline
					\textbf{Spectral range: }400-1000 nm
					\newline
					\textbf{GPU: }NVIDIA Tesla V100
					&
					\cite{qi2023field}
					
					\\
					\hline
					
					Y-Net
					&
					\textbf{Dataset}: 200 diseased corn leaves (extracted 6,264 regions)
					\newline
					\textbf{Type of disease:}  Brown spot and anthracnose
					\newline
					\textbf{Imaging Device} : An HSI system provided by Head Wall 
					\newline
					\textbf{Spectral range: }400-1000 nm
					\newline
					\textbf{GPU: }RTX 3090 24 Gb
					&
					\cite{jia2023net}
					
					\\
					\hline
				\end{tabular}

		\end{table*}
		
		The objective of neural network architecture design is to create a model that can effectively learn from the input data and generalize well to new, unseen data. In order to achieve this for HSI classification, a network architecture must be developed that can capture the complex spectral and spatial information present in the data. 
		
		As an integral aspect of designing a neural network, the decision of the number of layers to use is a crucial one.
		While the inclusion of more layers in a network has been shown \citep{uzair2020effects,josephine2021impact} to improve performance, it can simultaneously present several challenges. A notable challenge arises from the potential occurrence of vanishing or exploding gradients \citep{tan2019vanishing}, where the gradient signal becomes too small or too large as it backpropagates through the layers during training (see also Subsection \ref{hybrid_networks:resnet}). Such a phenomenon makes it difficult for the network to learn and adjust its weights effectively. 
		
		Moreover, as the number of layers and parameters increases, the risk of overfitting rises, which is characterized by the network's ability to perform exceptionally well on the training data, but not generalize well on unseen data. Consequently, the network may become computationally expensive to train and use due to its high processing power and memory requirements. Additionally, gradient computation time increases, thereby impeding the training's efficiency. Hence, careful consideration of these challenges and trade-offs is imperative to designing a network that balances complexity and performance.
		
		For this review,
		we classified 3D-CNNs into hybrid and non-hybrid structures. A network is hybrid if it includes either specific module(s) that improve feature extraction, accuracy, and performance or if it integrates 2D-CNN within the 3D-CNN architecture. In the following, we present 3D-CNN models for the classification of diseased  and defected crop using HSI data. A summary of these architectures is also given in
		Tables~\ref{table1} and~\ref{table2}.
		
		\subsection{Non-hybrid Networks} 
		
		\cite{nagasubramanian2018explaining} presented a 3D-CNN model to classify healthy and diseased crop. This model consists of two convolution layers with max pooling layers, and two Fully Connected (FC) layers, trained using the Adam optimizer. 
		To prevent overfitting, dropout mechanisms were used after the first max pooling and first FC layer. A Weighted Binary Cross Entropy (WBCE) function of the form 
		\begin{equation} \label{equ1}
			%		\begin{split}
				L_{WBCE}(y,\hat{y})=-[\beta \cdot y\log(\hat{y}) - (1-y)\log(1-\hat{y})]
				%		\end{split}
		\end{equation}
		was implemented to address imbalanced training data. Here $y$ and $\hat{y}$ represent binary variables for whether the ground truth and predicted result belong to a given class, respectively (see also Section \ref{aug} for a discussion on how to counteract imbalances in datasets). 
		Using the coefficient $\beta$ the WBCE loss function assigns higher weights to the minority class, for example, the false negatives rate decreases if $\beta$ is set higher than 1, while setting it smaller than 1 reduces the false positive rate. 
	
		Although the above technique can classify and detect diseased crop, detection of asymptomatic diseased crops at early stage and differentiating it from the healthy crops can be more challenging. In this respect, \cite{jung2022hyperspectral} developed a 3D-CNN model that improves classification accuracy of the asymptomatic diseased crop without modification and preprocessing of input HSI data. The model consists of four 3D convolution layers in which the first and fourth layers each are followed by a 3D max pooling layer and a batch normalization. The rest of the convolution layers are only followed by batch normalization. The output of the last 3D convolution layer is passed through a global average pooling and two dense layers.  In \cite{jung2022hyperspectral} the results were improved further by preprocessing the input HSI data which went through spectral differentiation, vertical expansion, and smoothing.
		
		Automating the classification procedure of
		HSI
		data using software can effectively mitigate the time-consuming process of implementing a deep learning pipeline. In this respect, \cite{gao2021hyperseed} developed an open-source software that classifies seeds at pixel level using 3D-CNN. Though at first it was developed for a specific crop (rice), the test experiments over other seeds are promising. This software utilizes a 3D-CNN that consists of two 3D convolution layers, with two and four 3D convolution kernels for the first and second layers, respectively. The output is then flattened via one FC layer and classification is performed using a softmax function. However, this software as part of its feature extraction process does not consider the global feature relationship which can improve its ability to recognize and classify seeds accurately.
		
		\cite{nguyen2021early} implemented a 3D-CNN for classification of a small size dataset. The feature extraction part is implemented based on AlexNet \citep{krizhevsky2017imagenet} and constitutes 5 convolutional layers followed by a flattening layer. The input layer takes an input of size 512$\times$512$\times$203, where 203 stands for number of bands. The first two convolutional layers
		are followed by a max pooling layer and a batch normalization.
		After that each convolutional layer is followed by a max pooling layer and after the last max pooling layer batch normalization and flattenings is applied before feeding the result into a RF or SVM for binary classification (healthy or diseased crop).
		
		\subsection{Hybrid networks}
		
		\subsubsection{3D-CNN architectures based on ResNet}\label{hybrid_networks:resnet}
		
		To address the challenges of vanishing or exploding gradients \cite{qiao2020detection} proposed to leverage residual convolutional blocks within a 3D deep ResNet architecture. This approach reduces the number of channels through the use of identity residual blocks and a convolutional residual blocks. 
		The identity residual block maintains the same input and output dimensions, while the convolutional residual block changes the number of channels. The use of a l$\times$l$\times$l convolution kernel as the shortcut in the convolutional residual block reduces the number of parameters and computational complexity.
		
		To further improve efficiency, \cite{qiao2020detection} adopt a bottleneck structure, that reduces the number of required convolution operations. Each convolutional layer is followed by a batch normalization layer to prevent vanishing gradient and enhance convergence rate. The network uses Exponential Linear Unit (ELU) as the non-linear activation function, which addresses the problem of dying neurons in ReLU.

		In order to identify the appropriate hyperparameters, the authors employed a Tree-structured Parzen Estimator (TPE) as an optimization algorithm. The TPE utilizes a probabilistic model to approximate the distribution of the objective function and guides the search for optimal hyperparameters.
		For the classification, a 7$\times$7$\times$1 global pooling layer and a FC are utilized. This model halves the number of parameters and improves the computational time up to 10\%. Moreover, the study of \cite{pourdarbani2023examination} over performance of well-known architectures in detection of defective crop demonstrates that residual connections achieve higher accuracy, while training faster despite having more parameters.
		
		\subsubsection{Hypernet-PRMF network}
		
		\cite{liu2020using} presented a feature pre-extraction and a multi-feature fusion block to extract peanut characteristics from hyperspectral data. The feature pre-extraction includes constructing a Peanut Recognition Index (PRI) based on two informative bands to distinguish healthy, moldy, and damaged peanuts. 
		
		The multi-feature fusion block is a technique used in image segmentation to fully extract spatial and spectral features from HSI data. This technique involves using multiple types of convolution kernels including 2D convolution for common texture features, separable convolution for increased feature diversity, depthwise convolution for band feature extraction, and 3D convolution for spectral change information. The convolutions are concatenated after normalization and activation functions to enhance diversity and improve recognition accuracy.
		
		Moreover, the authors employed feature pre-extraction and multi-feature fusion block techniques in their proposed peanut recognition model, called Hypernet-PRMF network. This model works at both 
		peanut- and pixel-level recognition. 
		The model consists of four parts: feature pre-extraction, down-sampling, up-sampling, and prediction.
		The feature pre-extraction part enhances the differentiation between different peanut features. The down-sampling part reduces the size of the image while increasing the number of convolution kernels, whereas the up-sampling part reconstructs the image while reducing the number of convolution kernels.
		The prediction works based on the softmax function and the class of maximum predicted probability is chosen as the final recognition result. The model achieves pixel-wise recognition accuracy with the use of the watershed segmentation algorithm.
		This technique has the potential to be employed for detection of other crops like sorghum.
		
		\subsubsection{Spectral Dilated Convolution 3D-CNN}
		
		\cite{cao2022detecting} proposed a spectral dilated convolution (SDC)-3D-CNN model to detect crop's asymptomatic diseases at an early stage. This model consists of SDC modules along with residual blocks that prevent the gradient vanishing problem. SDC extends the idea of dilated convolution which expands the receptive field of convolution kernels without augmenting the model's parameterization. Receptive field is the portion of the input space needed to create a filter at any convolutional layer. The 3D-SDC extends the receptive field of convolutional kernels to the spectral dimension. It works based on the principle of applying a filter to an input with intermittent intervals, which are dictated by the spectral dilation rate. 
	
		The network was tested with top 50, 100, 150, and 200 significant wavelengths extracted by RF and Principal Components (PCs) of the same ranking by PCA along with different spectral dilation rate to detect healthy, asymptomatic, and symptomatic crop. The experiment result shows higher detection performance of the network using top the 50 important features extracted by RF at a dilation rate of 5. 
		
		\subsubsection{Merged 2D- and 3D-CNN architectures}
		
		\cite{chen2022real} developed a 2D-3D-CNN for real-time crop defect detection. This network is the detection module of a real-time coffee-bean defect inspection algorithm (RT-CBDIA). The network consists of a 2D-CNN and a 3D-CNN, the former one is responsible to extract spatial features and the latter one is accountable for extraction of spectral features. Combining these two networks can boost feature extraction by providing robust and discriminative spectral-spatial features. 
	 
		The 3D-CNN is comprised of two convolution blocks with the same structure as in the 2D-CNN except that 3D convolutions and pooling layers are used.
		The two networks run simultaneously and their last pooling layers will be merged and fed to the a FC layer and then a dropout layer to avoid overfitting. Finally, a softmax layer determines each crop health status. 
		
		Likewise, \cite{qi2023field} fully extracted spatial-spectral features by merging 2D- and 3D-CNN architectures using  AttentionBlock \citep{yin2020novel} and Squeeze-and-Excitation (SE)-ResNet \citep{hu2018squeeze}. To accomplish this, the model first creates a neighborhood block of size 11$\times$11$\times$10 around a center pixel from the input image. Then, 2D convolution operations are used to extract spatial correlation features from the neighborhood block, and 3D convolution operations are used to capture spectral correlation features. The model uses four 2D convolutional layers and four 3D convolutional layers to capture feature maps of various spatial and spectral dimensions. By using different sizes of convolution kernels and downsampling steps, varied types of information can be captured. 
		
		Finally, the extracted feature maps are fused together to create a final set of feature maps  that contain valuable and pertinent information required for effective classification. Herein, AttentionBlock and SE-ResNet play important role, as outlined
		immediately below.
		
		An AttentionBlock is used to highlight important information in the fused spectral space feature map. It works by considering the similarity between each pixel in the feature map and weighting the relevant pixels with higher importance. This is achieved through a series of 2D convolutions with a kernel size of 1$\times$1 to transform each pixel into an $\lambda$-dimensional vector, where $\lambda$ represents the number of feature channels in the input tensor. The similarity between any two pixels is then calculated using the dot-product of their transformed vectors, and the results are weighted using a softmax function. 
		
		The output of an AttentionBlockt is a feature map that emphasizes the relevant information while suppressing irrelevant information. This process allows AttentionBlocks to focus on the relevance between pixels in the entire feature map, rather than just the spatial range of the convolution kernel size used in traditional convolution and pooling operations. This results in better classification results with little computational complexity.
		
		In order to enhance the representational power of CNNs, SE modules as a type of attention mechanism can adaptively recalibrate the feature maps. It does so by capturing channel-wise feature dependencies through a squeeze operation, followed by an excitation operation that learns how to weight the importance of each feature map. This mechanism allows the model to pay more attention to salient channel features and disregards the less significant ones. By integrating AttentionBlocks and SE-ResNet, the network of \cite{qi2023field} can better generalize in classification and achieve higher accuracy.
		
		Moreover, in the context of detection of two similar crop diseases that are indistinguishable to the naked eyes, a recent study by \cite{jia2023net} developed a new network called Y-Net. The Y-Net model takes in 10$\times$10$\times$203 hyperspectral data cubes as input and it consists of a channel attention mechanism, a band selection module with auxiliary classifier, a 3D-2D-CNN architecture, and a classification module.
		
		A CNN architecture is employed to conduct the band selection, where 1$\times$1 1D convolutions are assembled to modify the parameters of the convolutional kernel in the early phase of the network training. The magnitude of the weight of the convolution kernel is indicative of the relevance of the band, with greater absolute weight values signifying more distinctive bands. The use of group convolution helps in preventing any hindrance from nearby bands, while the ReLU activation function is adopted for fast network convergence without the problem of saturation. The output of this step will be given to an auxiliary classifier that enables early-stage weight updating in the band selection block.
		
		The auxiliary classifier module updates the loss function of the Y-Net model:
		\begin{multline}\label{equation2}
			L(y, \hat{y}) = (1 - \theta) \cdot L_{\text{Final Classifier}}(y, \hat{y}) \, + \\ \theta \cdot L_{\text{Auxiliary Classifier}}(y, \hat{y}) + \beta \cdot \sum_{j=1}^{n} W_j,
		\end{multline}
		where $y$ is the ground truth label, $\hat{y}$ is the predicted label, and $\theta$ and $\beta$ are hyperparameters that control the trade-off between the two losses and the sparsity of the band selection module, respectively. $W_j$  is the weight of the $j^{\text{th}}$ band in the band selection module and n is the number of total bands.
		The loss is a combination of the cross entropy losses of the final classifier and the auxiliary classifier and the sum of the weights of the band selection module. The purpose of this combination is to control the classification accuracy while updating the weights of the band selection layer. The adjustment factor $\theta$ gradually decreases as the number of training iterations increases. The presence of this adjustment factor enables the Y-Net model to update the weights in the band selection module in the early stages of training and gradually shift towards training the final classifier to learn a more accurate classification model. Additionally, the auxiliary classifier helps to constrain the weight sparsity of the band selection module, ensuring that the score of unimportant features is close to zero.
		
		The results of  \cite{jia2023net} show that by removing nonessential and nondiscriminative bands the accuracy of the Y-Net model increases and reduces the model size and the number of parameters. Moreover, since the band selection module is integrated into an overall architecture, the training time does not significantly increase.

		\section{Visualization techniques for HSI classification decisions} \label{visualization}
		
	Visualization techniques can be employed to observe the contribution of pixels in the classification decision. These techniques allow us to identify the pixel locations associated with the most important spectral bands that play a crucial role in the final classification results.
		
		Saliency maps \citep{simonyan2013deep} are one of the essential and traditional visualization tools to identify the most sensitive regions (crucial pixels) in an image with respect to a model's predictions. 
		This technique works by computing the gradient of the output class score with respect to the input image. This gradient represents how much each pixel in the input image contributes to the final classification decision. Next, the absolute values of these gradients are summed across the channels to obtain a saliency map, which highlights the most salient regions of the input image for the predicted class.
		
		For example, \cite{nagasubramanian2018explaining} discovered that saliency maps can help to locate the most sensitive pixel locations in infected crop images, which are often the severely infected areas. Conversely, both healthy and infected crop images had saliency map gradients that were primarily focused around the mid-region of the crop stem, highlighting the stem's importance in crop classification. Moreover, \cite{cao2022detecting} observed that the significant wavelengths extracted by RF from raw HSI data overlaps the saliency-sensitive wavelengths. More importantly, saliency maps can determine significant wavelengths for classification which are not extracted by RF.
		
		Another visual explanation technique for CNN decision is the Gradient-weighted Class Activation Mapping (Grad-CAM) \citep{selvaraju1610grad}. It was introduced as an improvement over CAM \citep{zhou2016learning}. 
		CAM involves modifying a pre-existing CNN model by replacing the final FC layer with a global average pooling layer, which retains essential channel information while reducing the spatial dimensions. This modification enables the utilization of feature maps from the preceding layer. By applying learned weights to these feature maps through global average pooling,
	CAM generates a map that highlights the crucial regions associated with the predicted category. 
		However, CAM provides a coarse localization of the important regions within an image. It highlights the regions that contribute most to the predicted class, but it does not provide precise boundaries of those regions. To address this limitation, Grad-CAM was introduced as an extension to CAM. 
		
		Grad-CAM enhances the CAM approach by incorporating gradient information. Similar to CAM, Grad-CAM also utilizes a global average pooling layer after the last convolutional layer to obtain importance scores for each channel in the feature maps. However, instead of learning separate linear models for each class, Grad-CAM calculates the gradients of the predicted class score with respect to the feature maps. These gradients are used to weigh the feature maps and enables Grad-CAM to identify more intricate features that play a significant role in the classification decision.
		
		In addition, there are further statistical techniques that help in analyzing the importance of individual features (spectral bands) of HSI data in a classification. Light Gradient Boosting Machine (LightGBM) \citep{ke2017lightgbm, gao2021hyperseed} is one of such methods that uses decision trees in calculating feature importance. It builds decision trees in a leaf-wise manner, meaning that the algorithm grows the tree by adding new leaves one at a time which is computationally faster compared to level-wise approach by selecting the best split based on the maximum reduction in loss function for all leaves in the tree. Therefore, LightGBM calculates the importance of each feature by evaluating how much it contributes to the reduction in loss function across all trees. 
		The feature importance score is calculated by summing up the number of times a feature is used to split the data across all trees, weighted by the improvement in accuracy achieved by each split. Features that are used more frequently and result in larger improvements in accuracy are assigned higher importance scores.
		
		\section{Discussion and conclusion}\label{discussion}

		In this study, we conducted a comprehensive review of 3D-CNN-based models applied in the domain of agriculture using non-UAV-based HSI data. Our analysis delved into diseased and defective crops, focusing on the structures and efficiencies of the models, the quantity of datasets utilized, and the necessary preprocessing steps. This review indicates the advantages of 3D-CNNs in capturing spatial-spectral information within hyperspectral data, enabling them to outperform 1D- and 2D-CNN in hyperspectral image classification.
		With the aim of assisting computer vision experts and agriculture-domain researchers in tackling HSI classification tasks for crops experiencing stress, this comprehensive review provides valuable insights and guidance.

		In general, HSI holds great potential for detecting subtle changes in crop growth and development, making it a promising technique for diagnosing crop diseases and defects. Despite this potential, our study indicates that there is still  limited research conducted that use 3D-CNNs in this context. Furthermore, the studies that are performed are often very application-specific and it is unknown how well the performed methods and models generalize to a broader range of applications. We identify three major challenges that must be overcome to achieve a broader adoption of HSI in general and the usage of 3D-CNN for HSI classification problems: limited availability of hyperspectral data, computational complexity of 3D-CNN models, and the costs of hyperspectral imaging hardware. In the following we address each of these challenges in more detail and offer ideas on how to overcome or avoid them.
		
		Limited availability of varied hyperspectral data on diseased and defective crops presents a significant challenge for researchers, farmers, and stakeholders in the agriculture industry who rely on data to make informed decisions. The lack of data in this area of the research hampers efforts to understand the extent of the problem and develop effective solutions. It makes it difficult to track the progress and success of any initiatives aimed at improving crop health and reducing the prevalence of disease and defects. To address this issue, there is a need for increased investment in data collection and data sharing, as it has been done in the past, for example, for RGB-data (e.g., \cite{Beck2020, Beck2022TheTC}). Large-scale HSI data collection and publicly available datasets will allow research groups to develop the next generation of models, even if they have no access to the otherwise required hardware and plant material. Even with small datasets there are also techniques to models beyond the specific application case they had been trained on. In recent years, transfer learning and active learning techniques have been increasingly used together to tackle the challenges posed by limited data. By leveraging knowledge from pre-existing models, transfer learning can enhance the accuracy of models trained on limited HSI data. On the other hand, active learning involves selecting and annotating informative samples to improve the effectiveness of models trained on small HSI datasets. By combining the two techniques, transfer learning and active learning can address the bottleneck of limited HSI data and enable the development of robust 3D-CNN models capable of accurately classifying HSI data.
		
		The computational complexity of 3D-CNN models is a barrier for their deployment in the field, for example, in edge computing devices or even as part of an embedded system in UAVs or agricultural equipment. Real-time diagnosis could significantly enable farmers to quickly detect and respond to disease outbreaks, which can help to prevent the spread of disease and reduce crop losses. This requires, however, researchers to explore ways to optimize their models for speed and efficiency, particularly their memory-footprint. This can be achieved through advancing the capabilities of single board computers, particularly their GPU and AI accelerators, on the one side, as well as developing lightweight models on the other side, for example, by reduction of used bands or grouping of bands into indices. Identifying the spectral bands that are most informative for detecting a varied range of diseases and defects would play an important role in reducing the model's training computational time, decreasing the number of parameters, achieving higher accuracy and generating a more lightweight model.		

		%NN: rewrote it as follows				
		Despite the valuable insights hyperspectral imaging provides regarding the condition and health of crops, deploying this technology is relatively costly compared to RGB and multispectral imaging technologies. The higher expense is primarily attributed to the increased processing power required to analyze the HSI data and the extensive range of spectrum offered by hyperspectral cameras.
		%CB: This paragraph sounds a bit dangerous to me.  At the begining of the paper, we argue for HSI over multi-spect.  Now at the end of the paper and all this work, it sounds like we're turning our back on HSI in favour of multi-spect.  Can this paragraph be rewritten to tone this down a bit?  NN: added the following lines

		One approach to consider for cost reduction is the limitation of the number of bands in hyperspectral cameras, as not all spectral bands may be equally crucial for disease and defect detection. 
		Therefore, instead of having an imaging device that supports full range of wavelengths from the visible to the infrared spectrum, we can deploy imaging systems that work with essential wavelengths rather than hundreds of spectral bands. In this respect, some companies already provide the facility to design customized multispectral systems, which work with bands that had been identified to be the most discriminative (see \cite{hamila2023fusarium} for a 3D-CNN model training advantage of this approach). Therefore, developing methods for identifying the most informative spectral bands would also be cost beneficial.
		Furthermore, by leveraging Machine Learning as a Service (MLaaS) platforms \citep{noshiri2021machine},
		the need for required infrastructure to process HSI data using 3D-CNN can be eliminated. Additionally, leading MLaaS providers can integrate pre-built 3D-CNN models into their offerings for training HSI data, thereby reducing the time and effort required for model development, especially for individuals in the agriculture domain who may have limited machine learning expertise.
		
		In summary, the application of 3D-CNNs with hyperspectral data for disease and defective crop detection is a promising research area with several open research questions. Collecting and sharing HSI data on scale, identifying informative spectral bands, developing transfer and active learning techniques, and implementing light-weight architectures are some of the key research areas that can be considered for future work. The advancement of this research will lead to more accurate and efficient disease and defective crop detection, which can have significant impacts on the agricultural industry.
		
		\section*{CRediT authorship contribution statement 
		}
	\textbf{Nooshin Noshiri}: Conceptualization, Investigation, Methodology, 
	Software, Visualization, Writing – original draft, Writing – review \& editing. \textbf{Michael A. Beck}: Writing - Original Draft, Writing – review \& editing. \textbf{Christopher P. Bidinosti}: Funding acquisition, Methodology, Project administration, Resources, Supervision, Writing - review \& editing. \textbf{Christopher J. Henry}: Conceptualization, Resources, Writing - review \& editing, Supervision, Project administration, Funding acquisition.
		\section*{Declaration of competing interest}
The authors declare that they have no known competing financial interests or personal relationships that could have appeared to influence the work reported in this paper.

		\section*{Acknowledgement}
		This research did not receive any specific grant from funding agencies in the public, commercial, or not-for-profit sectors.

		\bibliographystyle{elsarticle-harv} 
		\bibliography{ref}

\begin{thebibliography}{87}
\expandafter\ifx\csname natexlab\endcsname\relax\def\natexlab#1{#1}\fi
\providecommand{\url}[1]{\texttt{#1}}
\providecommand{\href}[2]{#2}
\providecommand{\path}[1]{#1}
\providecommand{\DOIprefix}{doi:}
\providecommand{\ArXivprefix}{arXiv:}
\providecommand{\URLprefix}{URL: }
\providecommand{\Pubmedprefix}{pmid:}
\providecommand{\doi}[1]{\href{http://dx.doi.org/#1}{\path{#1}}}
\providecommand{\Pubmed}[1]{\href{pmid:#1}{\path{#1}}}
\providecommand{\bibinfo}[2]{#2}
\ifx\xfnm\relax \def\xfnm[#1]{\unskip,\space#1}\fi
%Type = Article
\bibitem[{Avbelj(2012)}]{avbelj2012spectral}
\bibinfo{author}{Avbelj, J.}, \bibinfo{year}{2012}.
\newblock \bibinfo{title}{Spectral information retrieval for sub-pixel building
  edge detection}.
\newblock \bibinfo{journal}{ISPRS Annals of the Photogrammetry, Remote Sensing
  and Spatial Information Sciences} \bibinfo{volume}{I-7},
  \bibinfo{pages}{61--66}.
\newblock \DOIprefix\doi{https://doi.org/10.5194/isprsannals-I-7-61-2012}.
%Type = Article
\bibitem[{Beck et~al.(2022)Beck, Bidinosti, Henry and Ajmani}]{Beck2022TheTC}
\bibinfo{author}{Beck, M.A.}, \bibinfo{author}{Bidinosti, C.P.},
  \bibinfo{author}{Henry, C.J.}, \bibinfo{author}{Ajmani, M.},
  \bibinfo{year}{2022}.
\newblock \bibinfo{title}{The terrabyte client: providing access to terabytes
  of plant data}.
\newblock \bibinfo{journal}{ArXiv} \bibinfo{volume}{abs/2203.13691}.
\newblock \DOIprefix\doi{https://doi.org/10.48550/arXiv.2203.13691}.
%Type = Article
\bibitem[{Beck et~al.(2020)Beck, Liu, Bidinosti, Henry, Godee and
  Ajmani}]{Beck2020}
\bibinfo{author}{Beck, M.A.}, \bibinfo{author}{Liu, C.Y.},
  \bibinfo{author}{Bidinosti, C.P.}, \bibinfo{author}{Henry, C.J.},
  \bibinfo{author}{Godee, C.M.}, \bibinfo{author}{Ajmani, M.},
  \bibinfo{year}{2020}.
\newblock \bibinfo{title}{An embedded system for the automated generation of
  labeled plant images to enable machine learning applications in agriculture}.
\newblock \bibinfo{journal}{{PLOS} {ONE}} \bibinfo{volume}{15},
  \bibinfo{pages}{e0243923}.
\newblock \DOIprefix\doi{10.1371/journal.pone.0243923}.
%Type = Book
\bibitem[{Benediktsson and Ghamisi(2015)}]{benediktsson2015spectral}
\bibinfo{author}{Benediktsson, J.A.}, \bibinfo{author}{Ghamisi, P.},
  \bibinfo{year}{2015}.
\newblock \bibinfo{title}{Spectral-spatial classification of hyperspectral
  remote sensing images}.
\newblock \bibinfo{publisher}{Artech House}.
%Type = Inproceedings
\bibitem[{Bojeri et~al.(2022)Bojeri, Melgani, Giannotta, Ristorto, Guglieri and
  Junior}]{bojeri2022automatic}
\bibinfo{author}{Bojeri, A.}, \bibinfo{author}{Melgani, F.},
  \bibinfo{author}{Giannotta, G.}, \bibinfo{author}{Ristorto, G.},
  \bibinfo{author}{Guglieri, G.}, \bibinfo{author}{Junior, J.M.},
  \bibinfo{year}{2022}.
\newblock \bibinfo{title}{Automatic crop rows segmentation for multispectral
  aerial imagery}, in: \bibinfo{booktitle}{Mediterranean and Middle-East
  Geoscience and Remote Sensing Symposium (M2GARSS)},
  \bibinfo{organization}{IEEE}. pp. \bibinfo{pages}{78--81}.
\newblock \DOIprefix\doi{https://doi.org/10.1109/M2GARSS52314.2022.9839867}.
%Type = Article
\bibitem[{Boston et~al.(2022)Boston, Van~Dijk, Larraondo and
  Thackway}]{boston2022comparing}
\bibinfo{author}{Boston, T.}, \bibinfo{author}{Van~Dijk, A.},
  \bibinfo{author}{Larraondo, P.R.}, \bibinfo{author}{Thackway, R.},
  \bibinfo{year}{2022}.
\newblock \bibinfo{title}{Comparing cnns and random forests for landsat image
  segmentation trained on a large proxy land cover dataset}.
\newblock \bibinfo{journal}{Remote Sensing} \bibinfo{volume}{14},
  \bibinfo{pages}{3396}.
\newblock \DOIprefix\doi{https://doi.org/10.3390/rs14143396}.
%Type = Article
\bibitem[{Bunting et~al.(2014)Bunting, Clewley, Lucas and
  Gillingham}]{Bunting_etal_2014}
\bibinfo{author}{Bunting, P.}, \bibinfo{author}{Clewley, D.},
  \bibinfo{author}{Lucas, R.M.}, \bibinfo{author}{Gillingham, S.},
  \bibinfo{year}{2014}.
\newblock \bibinfo{title}{{The Remote Sensing and GIS Software Library
  (RSGISLib)}}.
\newblock \bibinfo{journal}{Computers and Geosciences} \bibinfo{volume}{62},
  \bibinfo{pages}{216--226}.
\newblock \DOIprefix\doi{https://doi.org/10.1016/j.cageo.2013.08.007}.
%Type = Article
\bibitem[{Cao et~al.(2017)Cao, Ji, Ji, Wang and Jiao}]{cao2017hyperspectral}
\bibinfo{author}{Cao, X.}, \bibinfo{author}{Ji, B.}, \bibinfo{author}{Ji, Y.},
  \bibinfo{author}{Wang, L.}, \bibinfo{author}{Jiao, L.}, \bibinfo{year}{2017}.
\newblock \bibinfo{title}{Hyperspectral image classification based on
  filtering: a comparative study}.
\newblock \bibinfo{journal}{Journal of Applied Remote Sensing}
  \bibinfo{volume}{11}, \bibinfo{pages}{035007--035007}.
\newblock \DOIprefix\doi{https://doi.org/10.1117/1.JRS.11.035007}.
%Type = Article
\bibitem[{Cao et~al.(2022)Cao, Yuan, Xu, Mart{\'\i}nez-Ortega, Feng and
  Zhai}]{cao2022detecting}
\bibinfo{author}{Cao, Y.}, \bibinfo{author}{Yuan, P.}, \bibinfo{author}{Xu,
  H.}, \bibinfo{author}{Mart{\'\i}nez-Ortega, J.F.}, \bibinfo{author}{Feng,
  J.}, \bibinfo{author}{Zhai, Z.}, \bibinfo{year}{2022}.
\newblock \bibinfo{title}{Detecting asymptomatic infections of rice bacterial
  leaf blight using hyperspectral imaging and 3-dimensional convolutional
  neural network with spectral dilated convolution}.
\newblock \bibinfo{journal}{Frontiers in Plant Science} \bibinfo{volume}{13}.
\newblock \DOIprefix\doi{https://doi.org/10.3389/fpls.2022.963170}.
%Type = Misc
\bibitem[{{Centre National D'Etudes Spatiales (CNES)}(2023)}]{CNES}
\bibinfo{author}{{Centre National D'Etudes Spatiales (CNES)}},
  \bibinfo{year}{2023}.
\newblock \bibinfo{title}{Orfeo toolbox}.
\newblock \URLprefix \url{https://www.orfeo-toolbox.org/download/}.
%Type = Article
\bibitem[{Chawla et~al.(2002)Chawla, Bowyer, Hall and
  Kegelmeyer}]{chawla2002smote}
\bibinfo{author}{Chawla, N.V.}, \bibinfo{author}{Bowyer, K.W.},
  \bibinfo{author}{Hall, L.O.}, \bibinfo{author}{Kegelmeyer, W.P.},
  \bibinfo{year}{2002}.
\newblock \bibinfo{title}{Smote: synthetic minority over-sampling technique}.
\newblock \bibinfo{journal}{Journal of artificial intelligence research}
  \bibinfo{volume}{16}, \bibinfo{pages}{321--357}.
\newblock \DOIprefix\doi{https://doi.org/10.1613/jair.953}.
%Type = Article
\bibitem[{Chen et~al.(2022)Chen, Chiu and Zou}]{chen2022real}
\bibinfo{author}{Chen, S.Y.}, \bibinfo{author}{Chiu, M.F.},
  \bibinfo{author}{Zou, X.W.}, \bibinfo{year}{2022}.
\newblock \bibinfo{title}{Real-time defect inspection of green coffee beans
  using nir snapshot hyperspectral imaging}.
\newblock \bibinfo{journal}{Computers and Electronics in Agriculture}
  \bibinfo{volume}{197}, \bibinfo{pages}{106970}.
\newblock \DOIprefix\doi{https://doi.org/10.1016/j.compag.2022.106970}.
%Type = Article
\bibitem[{Dalla~Mura et~al.(2010)Dalla~Mura, Benediktsson, Waske and
  Bruzzone}]{dalla2010morphological}
\bibinfo{author}{Dalla~Mura, M.}, \bibinfo{author}{Benediktsson, J.A.},
  \bibinfo{author}{Waske, B.}, \bibinfo{author}{Bruzzone, L.},
  \bibinfo{year}{2010}.
\newblock \bibinfo{title}{Morphological attribute profiles for the analysis of
  very high resolution images}.
\newblock \bibinfo{journal}{IEEE Transactions on Geoscience and Remote Sensing}
  \bibinfo{volume}{48}, \bibinfo{pages}{3747--3762}.
\newblock \DOIprefix\doi{https://doi.org/10.1109/TGRS.2010.2048116}.
%Type = Article
\bibitem[{Damodaran et~al.(2017)Damodaran, Courty and
  Lef{\`e}vre}]{damodaran2017sparse}
\bibinfo{author}{Damodaran, B.B.}, \bibinfo{author}{Courty, N.},
  \bibinfo{author}{Lef{\`e}vre, S.}, \bibinfo{year}{2017}.
\newblock \bibinfo{title}{Sparse hilbert schmidt independence criterion and
  surrogate-kernel-based feature selection for hyperspectral image
  classification}.
\newblock \bibinfo{journal}{IEEE Transactions on Geoscience and Remote Sensing}
  \bibinfo{volume}{55}, \bibinfo{pages}{2385--2398}.
\newblock \DOIprefix\doi{https://doi.org/10.1109/TGRS.2016.2642479}.
%Type = Article
\bibitem[{Datta et~al.(2015)Datta, Ghosh and Ghosh}]{datta2015combination}
\bibinfo{author}{Datta, A.}, \bibinfo{author}{Ghosh, S.},
  \bibinfo{author}{Ghosh, A.}, \bibinfo{year}{2015}.
\newblock \bibinfo{title}{Combination of clustering and ranking techniques for
  unsupervised band selection of hyperspectral images}.
\newblock \bibinfo{journal}{IEEE Journal of Selected Topics in Applied Earth
  Observations and Remote Sensing} \bibinfo{volume}{8},
  \bibinfo{pages}{2814--2823}.
\newblock \DOIprefix\doi{https://doi.org/10.1109/JSTARS.2015.2428276}.
%Type = Book
\bibitem[{Duggal(2013)}]{duggal2013surveying}
\bibinfo{author}{Duggal, S.}, \bibinfo{year}{2013}.
\newblock \bibinfo{title}{Surveying}. volume~\bibinfo{volume}{2}.
\newblock \bibinfo{publisher}{Tata McGraw-Hill Education}.
%Type = Article
\bibitem[{F{\i}rat et~al.(2022)F{\i}rat, Asker and
  Hanbay}]{firat2022classification}
\bibinfo{author}{F{\i}rat, H.}, \bibinfo{author}{Asker, M.E.},
  \bibinfo{author}{Hanbay, D.}, \bibinfo{year}{2022}.
\newblock \bibinfo{title}{Classification of hyperspectral remote sensing images
  using different dimension reduction methods with 3d/2d cnn}.
\newblock \bibinfo{journal}{Remote Sensing Applications: Society and
  Environment} \bibinfo{volume}{25}, \bibinfo{pages}{100694}.
\newblock \DOIprefix\doi{https://doi.org/10.1016/j.rsase.2022.100694}.
%Type = Article
\bibitem[{Gao et~al.(2021)Gao, Chandran, Paul, Walia and Yu}]{gao2021hyperseed}
\bibinfo{author}{Gao, T.}, \bibinfo{author}{Chandran, A.K.N.},
  \bibinfo{author}{Paul, P.}, \bibinfo{author}{Walia, H.}, \bibinfo{author}{Yu,
  H.}, \bibinfo{year}{2021}.
\newblock \bibinfo{title}{Hyperseed: An end-to-end method to process
  hyperspectral images of seeds}.
\newblock \bibinfo{journal}{Sensors} \bibinfo{volume}{21},
  \bibinfo{pages}{8184}.
\newblock \DOIprefix\doi{https://doi.org/10.3390/s21248184}.
%Type = Article
\bibitem[{Ghamisi et~al.(2014)Ghamisi, Couceiro and
  Benediktsson}]{ghamisi2014novel}
\bibinfo{author}{Ghamisi, P.}, \bibinfo{author}{Couceiro, M.S.},
  \bibinfo{author}{Benediktsson, J.A.}, \bibinfo{year}{2014}.
\newblock \bibinfo{title}{A novel feature selection approach based on fodpso
  and svm}.
\newblock \bibinfo{journal}{IEEE Transactions on Geoscience and Remote Sensing}
  \bibinfo{volume}{53}, \bibinfo{pages}{2935--2947}.
\newblock \DOIprefix\doi{https://doi.org/10.1109/TGRS.2014.2367010}.
%Type = Article
\bibitem[{Gill et~al.(2022)Gill, Gill, Saini, Chopra, de~Koff and
  Sandhu}]{gill2022comprehensive}
\bibinfo{author}{Gill, T.}, \bibinfo{author}{Gill, S.K.},
  \bibinfo{author}{Saini, D.K.}, \bibinfo{author}{Chopra, Y.},
  \bibinfo{author}{de~Koff, J.P.}, \bibinfo{author}{Sandhu, K.S.},
  \bibinfo{year}{2022}.
\newblock \bibinfo{title}{A comprehensive review of high throughput phenotyping
  and machine learning for plant stress phenotyping}.
\newblock \bibinfo{journal}{Phenomics} \bibinfo{volume}{2},
  \bibinfo{pages}{156--183}.
\newblock \DOIprefix\doi{https://doi.org/10.1007/s43657-022-00048-z}.
%Type = Article
\bibitem[{Hamila et~al.(2023)Hamila, Henry, Molina, Bidinosti and
  Henriquez}]{hamila2023fusarium}
\bibinfo{author}{Hamila, O.}, \bibinfo{author}{Henry, C.J.},
  \bibinfo{author}{Molina, O.I.}, \bibinfo{author}{Bidinosti, C.P.},
  \bibinfo{author}{Henriquez, M.A.}, \bibinfo{year}{2023}.
\newblock \bibinfo{title}{Fusarium head blight detection, spikelet estimation,
  and severity assessment in wheat using 3d convolutional neural networks}.
\newblock \bibinfo{journal}{arXiv} \bibinfo{volume}{abs/2303.05634}.
\newblock \DOIprefix\doi{https://doi.org/10.3390/rs11222658}.
%Type = Inproceedings
\bibitem[{He et~al.(2008)He, Bai, Garcia and Li}]{he2008adasyn}
\bibinfo{author}{He, H.}, \bibinfo{author}{Bai, Y.}, \bibinfo{author}{Garcia,
  E.A.}, \bibinfo{author}{Li, S.}, \bibinfo{year}{2008}.
\newblock \bibinfo{title}{Adasyn: Adaptive synthetic sampling approach for
  imbalanced learning}, in: \bibinfo{booktitle}{Proceedings of International
  Joint Conference on Neural Networks (IJCNN)}, \bibinfo{publisher}{IEEE},
  \bibinfo{address}{Hong Kong}. pp. \bibinfo{pages}{1322--1328}.
\newblock \DOIprefix\doi{https://doi.org/10.1109/IJCNN.2008.4633969}.
%Type = Inproceedings
\bibitem[{Henrich et~al.(2009)Henrich, G{\"o}tze, Jung, Sandow, Th{\"u}rkow and
  Gl{\"a}{\ss}er}]{henrich2009development}
\bibinfo{author}{Henrich, V.}, \bibinfo{author}{G{\"o}tze, E.},
  \bibinfo{author}{Jung, A.}, \bibinfo{author}{Sandow, C.},
  \bibinfo{author}{Th{\"u}rkow, D.}, \bibinfo{author}{Gl{\"a}{\ss}er, C.},
  \bibinfo{year}{2009}.
\newblock \bibinfo{title}{Development of an online indices database:
  Motivation, concept and implementation}, in: \bibinfo{booktitle}{6th EARSeL
  imaging spectroscopy sig workshop innovative tool for scientific and
  commercial environment applications}, \bibinfo{address}{Tel Aviv, Israel}.
  pp. \bibinfo{pages}{16--18}.
\newblock \URLprefix \url{https://www.indexdatabase.de/info/credits.php}.
%Type = Inproceedings
\bibitem[{Hu et~al.(2018)Hu, Shen and Sun}]{hu2018squeeze}
\bibinfo{author}{Hu, J.}, \bibinfo{author}{Shen, L.}, \bibinfo{author}{Sun,
  G.}, \bibinfo{year}{2018}.
\newblock \bibinfo{title}{Squeeze-and-excitation networks}, in:
  \bibinfo{booktitle}{Proceedings of Computer Vision and Pattern Recognition
  Conference (CVPR)}, \bibinfo{publisher}{IEEE}, \bibinfo{address}{Salt Lake
  City, UT, USA}. pp. \bibinfo{pages}{7132--7141}.
\newblock \DOIprefix\doi{https://doi.org/10.1109/CVPR.2018.00745}.
%Type = Article
\bibitem[{Ifarraguerri and Prairie(2004)}]{ifarraguerri2004visual}
\bibinfo{author}{Ifarraguerri, A.}, \bibinfo{author}{Prairie, M.W.},
  \bibinfo{year}{2004}.
\newblock \bibinfo{title}{Visual method for spectral band selection}.
\newblock \bibinfo{journal}{IEEE Geoscience and Remote Sensing Letters}
  \bibinfo{volume}{1}, \bibinfo{pages}{101--106}.
\newblock \DOIprefix\doi{https://doi.org/10.1109/LGRS.2003.822879}.
%Type = Inproceedings
\bibitem[{Imbiriba et~al.(2015)Imbiriba, Bermudez, Richard and
  Tourneret}]{imbiriba2015band}
\bibinfo{author}{Imbiriba, T.}, \bibinfo{author}{Bermudez, J.C.M.},
  \bibinfo{author}{Richard, C.}, \bibinfo{author}{Tourneret, J.Y.},
  \bibinfo{year}{2015}.
\newblock \bibinfo{title}{Band selection in rkhs for fast nonlinear unmixing of
  hyperspectral images}, in: \bibinfo{booktitle}{Proceedings of European Signal
  Processing Conference (EUSIPCO)}, \bibinfo{publisher}{IEEE},
  \bibinfo{address}{Nice, France}. pp. \bibinfo{pages}{1651--1655}.
\newblock \DOIprefix\doi{https://doi.org/10.1109/EUSIPCO.2015.7362664}.
%Type = Article
\bibitem[{Jia et~al.(2023)Jia, Shi, Luo and Sun}]{jia2023net}
\bibinfo{author}{Jia, Y.}, \bibinfo{author}{Shi, Y.}, \bibinfo{author}{Luo,
  J.}, \bibinfo{author}{Sun, H.}, \bibinfo{year}{2023}.
\newblock \bibinfo{title}{Y--net: Identification of typical diseases of corn
  leaves using a 3d--2d hybrid cnn model combined with a hyperspectral image
  band selection module}.
\newblock \bibinfo{journal}{Sensors} \bibinfo{volume}{23},
  \bibinfo{pages}{1494}.
\newblock \DOIprefix\doi{https://doi.org/10.3390/s23031494}.
%Type = Article
\bibitem[{Jiang and Li(2020)}]{jiang2020convolutional}
\bibinfo{author}{Jiang, Y.}, \bibinfo{author}{Li, C.}, \bibinfo{year}{2020}.
\newblock \bibinfo{title}{Convolutional neural networks for image-based
  high-throughput plant phenotyping: a review}.
\newblock \bibinfo{journal}{Plant Phenomics} \bibinfo{volume}{2020}.
\newblock \DOIprefix\doi{https://doi.org/10.34133/2020/4152816}.
%Type = Article
\bibitem[{Jin et~al.(2018)Jin, Jie, Wang, Qi and Li}]{jin2018classifying}
\bibinfo{author}{Jin, X.}, \bibinfo{author}{Jie, L.}, \bibinfo{author}{Wang,
  S.}, \bibinfo{author}{Qi, H.J.}, \bibinfo{author}{Li, S.W.},
  \bibinfo{year}{2018}.
\newblock \bibinfo{title}{Classifying wheat hyperspectral pixels of healthy
  heads and fusarium head blight disease using a deep neural network in the
  wild field}.
\newblock \bibinfo{journal}{Remote Sensing} \bibinfo{volume}{10},
  \bibinfo{pages}{395}.
\newblock \DOIprefix\doi{https://doi.org/10.3390/rs10030395}.
%Type = Article
\bibitem[{Joseph et~al.(2022)Joseph, Pawar and
  Pramanik}]{joseph2022intelligent}
\bibinfo{author}{Joseph, D.S.}, \bibinfo{author}{Pawar, P.M.},
  \bibinfo{author}{Pramanik, R.}, \bibinfo{year}{2022}.
\newblock \bibinfo{title}{Intelligent plant disease diagnosis using
  convolutional neural network: a review}.
\newblock \bibinfo{journal}{Multimedia Tools and Applications}
  \bibinfo{volume}{82}, \bibinfo{pages}{1--67}.
\newblock \DOIprefix\doi{http://dx.doi.org/10.1007/s11042-022-14004-6}.
%Type = Inproceedings
\bibitem[{Josephine et~al.(2021)Josephine, Nirmala and
  Alluri}]{josephine2021impact}
\bibinfo{author}{Josephine, V.H.}, \bibinfo{author}{Nirmala, A.},
  \bibinfo{author}{Alluri, V.L.}, \bibinfo{year}{2021}.
\newblock \bibinfo{title}{Impact of hidden dense layers in convolutional neural
  network to enhance performance of classification model}, in:
  \bibinfo{booktitle}{IOP Conference Series: Materials Science and
  Engineering}, \bibinfo{organization}{IOP Publishing}. p.
  \bibinfo{pages}{012007}.
\newblock \DOIprefix\doi{https://doi.org/10.1088/1757-899X/1131/1/012007}.
%Type = Article
\bibitem[{Jung et~al.(2022)Jung, Kim, Kim, Lee, Kim and
  Park}]{jung2022hyperspectral}
\bibinfo{author}{Jung, D.H.}, \bibinfo{author}{Kim, J.D.},
  \bibinfo{author}{Kim, H.Y.}, \bibinfo{author}{Lee, T.S.},
  \bibinfo{author}{Kim, H.S.}, \bibinfo{author}{Park, S.},
  \bibinfo{year}{2022}.
\newblock \bibinfo{title}{A hyperspectral data 3d convolutional neural network
  classification model for diagnosis of gray mold disease in strawberry
  leaves}.
\newblock \bibinfo{journal}{Frontiers in Plant Science} \bibinfo{volume}{13},
  \bibinfo{pages}{620}.
\newblock \DOIprefix\doi{https://doi.org/10.3389/fpls.2022.837020}.
%Type = Article
\bibitem[{Ke et~al.(2017)Ke, Meng, Finley, Wang, Chen, Ma, Ye and
  Liu}]{ke2017lightgbm}
\bibinfo{author}{Ke, G.}, \bibinfo{author}{Meng, Q.}, \bibinfo{author}{Finley,
  T.}, \bibinfo{author}{Wang, T.}, \bibinfo{author}{Chen, W.},
  \bibinfo{author}{Ma, W.}, \bibinfo{author}{Ye, Q.}, \bibinfo{author}{Liu,
  T.Y.}, \bibinfo{year}{2017}.
\newblock \bibinfo{title}{Lightgbm: A highly efficient gradient boosting
  decision tree}.
\newblock \bibinfo{journal}{Advances in neural information processing systems}
  \bibinfo{volume}{30}.
%Type = Inproceedings
\bibitem[{Khodr and Younes(2011)}]{khodr2011dimensionality}
\bibinfo{author}{Khodr, J.}, \bibinfo{author}{Younes, R.},
  \bibinfo{year}{2011}.
\newblock \bibinfo{title}{Dimensionality reduction on hyperspectral images: A
  comparative review based on artificial datas}, in:
  \bibinfo{booktitle}{Proceedings of International Congress on Image and Signal
  Processing}, \bibinfo{publisher}{IEEE}, \bibinfo{address}{Shanghai, China}.
  pp. \bibinfo{pages}{1875--1883}.
\newblock \DOIprefix\doi{https://doi.org/10.1109/CISP.2011.6100531}.
%Type = Article
\bibitem[{Krizhevsky et~al.(2017)Krizhevsky, Sutskever and
  Hinton}]{krizhevsky2017imagenet}
\bibinfo{author}{Krizhevsky, A.}, \bibinfo{author}{Sutskever, I.},
  \bibinfo{author}{Hinton, G.E.}, \bibinfo{year}{2017}.
\newblock \bibinfo{title}{Imagenet classification with deep convolutional
  neural networks}.
\newblock \bibinfo{journal}{Communications of the ACM} \bibinfo{volume}{60},
  \bibinfo{pages}{84--90}.
\newblock \DOIprefix\doi{http://dx.doi.org/10.1145/3065386}.
%Type = Article
\bibitem[{Kumar et~al.(2015)Kumar, Gupta, Mishra and
  Prasad}]{kumar2015comparison}
\bibinfo{author}{Kumar, P.}, \bibinfo{author}{Gupta, D.K.},
  \bibinfo{author}{Mishra, V.N.}, \bibinfo{author}{Prasad, R.},
  \bibinfo{year}{2015}.
\newblock \bibinfo{title}{Comparison of support vector machine, artificial
  neural network, and spectral angle mapper algorithms for crop classification
  using liss iv data}.
\newblock \bibinfo{journal}{International Journal of Remote Sensing}
  \bibinfo{volume}{36}, \bibinfo{pages}{1604--1617}.
\newblock \DOIprefix\doi{https://doi.org/10.1080/2150704X.2015.1019015}.
%Type = Misc
\bibitem[{{L3Harris Geospatial}(2023)}]{HG}
\bibinfo{author}{{L3Harris Geospatial}}, \bibinfo{year}{2023}.
\newblock \bibinfo{title}{Envi}.
\newblock \URLprefix
  \url{https://www.l3harrisgeospatial.com/Software-Technology/ENVI}.
%Type = Inproceedings
\bibitem[{Lennon et~al.(2002)Lennon, Mercier and
  Hubert-Moy}]{lennon2002nonlinear}
\bibinfo{author}{Lennon, M.}, \bibinfo{author}{Mercier, G.},
  \bibinfo{author}{Hubert-Moy, L.}, \bibinfo{year}{2002}.
\newblock \bibinfo{title}{Nonlinear filtering of hyperspectral images with
  anisotropic diffusion}, in: \bibinfo{booktitle}{Proceedings of International
  Geoscience and Remote Sensing Symposium}, \bibinfo{publisher}{IEEE},
  \bibinfo{address}{Toronto, ON, Canada}. pp. \bibinfo{pages}{2477--2479}.
\newblock \DOIprefix\doi{https://doi.org/10.1109/IGARSS.2002.1026583}.
%Type = Article
\bibitem[{Li et~al.(2022)Li, Wang, Zhang and Liu}]{li2022semantic}
\bibinfo{author}{Li, J.}, \bibinfo{author}{Wang, H.}, \bibinfo{author}{Zhang,
  A.}, \bibinfo{author}{Liu, Y.}, \bibinfo{year}{2022}.
\newblock \bibinfo{title}{Semantic segmentation of hyperspectral remote sensing
  images based on pse-unet model}.
\newblock \bibinfo{journal}{Sensors} \bibinfo{volume}{22},
  \bibinfo{pages}{9678}.
\newblock \DOIprefix\doi{https://doi.org/10.3390/s22249678}.
%Type = Article
\bibitem[{Li and Qian(2011)}]{li2011clustering}
\bibinfo{author}{Li, J.m.}, \bibinfo{author}{Qian, Y.t.}, \bibinfo{year}{2011}.
\newblock \bibinfo{title}{Clustering-based hyperspectral band selection using
  sparse nonnegative matrix factorization}.
\newblock \bibinfo{journal}{Journal of Zhejiang University Science C}
  \bibinfo{volume}{12}, \bibinfo{pages}{542--549}.
\newblock \DOIprefix\doi{https://doi.org/10.1631/jzus.C1000304}.
%Type = Article
\bibitem[{Liu et~al.(2020)Liu, Jiang, Qiao, Qi, Pan and Pan}]{liu2020using}
\bibinfo{author}{Liu, Z.}, \bibinfo{author}{Jiang, J.}, \bibinfo{author}{Qiao,
  X.}, \bibinfo{author}{Qi, X.}, \bibinfo{author}{Pan, Y.},
  \bibinfo{author}{Pan, X.}, \bibinfo{year}{2020}.
\newblock \bibinfo{title}{Using convolution neural network and hyperspectral
  image to identify moldy peanut kernels}.
\newblock \bibinfo{journal}{LWT} \bibinfo{volume}{132},
  \bibinfo{pages}{109815}.
\newblock \DOIprefix\doi{https://doi.org/10.1016/j.lwt.2020.109815}.
%Type = Article
\bibitem[{Luo et~al.(2016)Luo, Chen, Tian, Qin and Qian}]{luo2016minimum}
\bibinfo{author}{Luo, G.}, \bibinfo{author}{Chen, G.}, \bibinfo{author}{Tian,
  L.}, \bibinfo{author}{Qin, K.}, \bibinfo{author}{Qian, S.E.},
  \bibinfo{year}{2016}.
\newblock \bibinfo{title}{Minimum noise fraction versus principal component
  analysis as a preprocessing step for hyperspectral imagery denoising}.
\newblock \bibinfo{journal}{Canadian Journal of Remote Sensing}
  \bibinfo{volume}{42}, \bibinfo{pages}{106--116}.
\newblock \DOIprefix\doi{https://doi.org/10.1080/07038992.2016.1160772}.
%Type = Article
\bibitem[{Mojaradi et~al.(2008)Mojaradi, Emami, Varshosaz and
  Jamali}]{mojaradib2008novelband}
\bibinfo{author}{Mojaradi, B.}, \bibinfo{author}{Emami, H.},
  \bibinfo{author}{Varshosaz, M.}, \bibinfo{author}{Jamali, S.},
  \bibinfo{year}{2008}.
\newblock \bibinfo{title}{A novelband selection methodfor hyperspectral data
  analysis}.
\newblock \bibinfo{journal}{Proceedings of International Archives of
  Photogrammry, Remote Sensing and Spatial Information Sciences}
  \bibinfo{volume}{447}, \bibinfo{pages}{451}.
%Type = Article
\bibitem[{Nagasubramanian et~al.(2018)Nagasubramanian, Jones, Singh, Singh,
  Ganapathysubramanian and Sarkar}]{nagasubramanian2018explaining}
\bibinfo{author}{Nagasubramanian, K.}, \bibinfo{author}{Jones, S.},
  \bibinfo{author}{Singh, A.K.}, \bibinfo{author}{Singh, A.},
  \bibinfo{author}{Ganapathysubramanian, B.}, \bibinfo{author}{Sarkar, S.},
  \bibinfo{year}{2018}.
\newblock \bibinfo{title}{Explaining hyperspectral imaging based plant disease
  identification: 3d cnn and saliency maps}.
\newblock \bibinfo{journal}{arXiv} \bibinfo{volume}{abs/1804.08831}.
%Type = Misc
\bibitem[{{NASA Ocean Biology Processing Group (OBPG)}(2022)}]{NASA}
\bibinfo{author}{{NASA Ocean Biology Processing Group (OBPG)}},
  \bibinfo{year}{2022}.
\newblock \bibinfo{title}{Seadas}.
\newblock \URLprefix \url{https://seadas.gsfc.nasa.gov/downloads/}.
%Type = Article
\bibitem[{Nguyen et~al.(2021)Nguyen, Sagan, Maimaitiyiming, Maimaitijiang,
  Bhadra and Kwasniewski}]{nguyen2021early}
\bibinfo{author}{Nguyen, C.}, \bibinfo{author}{Sagan, V.},
  \bibinfo{author}{Maimaitiyiming, M.}, \bibinfo{author}{Maimaitijiang, M.},
  \bibinfo{author}{Bhadra, S.}, \bibinfo{author}{Kwasniewski, M.T.},
  \bibinfo{year}{2021}.
\newblock \bibinfo{title}{Early detection of plant viral disease using
  hyperspectral imaging and deep learning}.
\newblock \bibinfo{journal}{Sensors} \bibinfo{volume}{21},
  \bibinfo{pages}{742}.
\newblock \DOIprefix\doi{https://doi.org/10.3390/s21030742}.
%Type = Inproceedings
\bibitem[{Noshiri et~al.(2021)Noshiri, Khorramfar and
  Halabi}]{noshiri2021machine}
\bibinfo{author}{Noshiri, N.}, \bibinfo{author}{Khorramfar, M.},
  \bibinfo{author}{Halabi, T.}, \bibinfo{year}{2021}.
\newblock \bibinfo{title}{Machine learning-as-a-service performance evaluation
  on multi-class datasets}, in: \bibinfo{booktitle}{International Conference on
  Smart Internet of Things (SmartIoT)}, \bibinfo{organization}{IEEE}. pp.
  \bibinfo{pages}{332--336}.
\newblock \DOIprefix\doi{https://doi.org/10.1109/SmartIoT52359.2021.00060}.
%Type = Inbook
\bibitem[{Plaza et~al.(2011)Plaza, Mart{\'i}n, Plaza, Zortea and
  S{\'a}nchez}]{plaza2011recent}
\bibinfo{author}{Plaza, A.}, \bibinfo{author}{Mart{\'i}n, G.},
  \bibinfo{author}{Plaza, J.}, \bibinfo{author}{Zortea, M.},
  \bibinfo{author}{S{\'a}nchez, S.}, \bibinfo{year}{2011}.
\newblock \bibinfo{title}{Recent Developments in Endmember Extraction and
  Spectral Unmixing}. \bibinfo{publisher}{Springer Berlin Heidelberg}.
\newblock pp. \bibinfo{pages}{235--267}.
\newblock \DOIprefix\doi{https://doi.org/10.1007/978-3-642-14212-3_12}.
%Type = Article
\bibitem[{Pourdarbani et~al.(2023)Pourdarbani, Sabzi, Dehghankar, Rohban and
  Arribas}]{pourdarbani2023examination}
\bibinfo{author}{Pourdarbani, R.}, \bibinfo{author}{Sabzi, S.},
  \bibinfo{author}{Dehghankar, M.}, \bibinfo{author}{Rohban, M.H.},
  \bibinfo{author}{Arribas, J.I.}, \bibinfo{year}{2023}.
\newblock \bibinfo{title}{Examination of lemon bruising using different
  cnn-based classifiers and local spectral-spatial hyperspectral imaging}.
\newblock \bibinfo{journal}{Algorithms} \bibinfo{volume}{16},
  \bibinfo{pages}{113}.
\newblock \DOIprefix\doi{https://doi.org/10.3390/a16020113}.
%Type = Article
\bibitem[{Qi et~al.(2023)Qi, Sandroni, Westergaard, Sundmark, Bagge,
  Alexandersson and Gao}]{qi2023field}
\bibinfo{author}{Qi, C.}, \bibinfo{author}{Sandroni, M.},
  \bibinfo{author}{Westergaard, J.C.}, \bibinfo{author}{Sundmark, E.H.R.},
  \bibinfo{author}{Bagge, M.}, \bibinfo{author}{Alexandersson, E.},
  \bibinfo{author}{Gao, J.}, \bibinfo{year}{2023}.
\newblock \bibinfo{title}{In-field classification of the asymptomatic
  biotrophic phase of potato late blight based on deep learning and proximal
  hyperspectral imaging}.
\newblock \bibinfo{journal}{Computers and Electronics in Agriculture}
  \bibinfo{volume}{205}, \bibinfo{pages}{107585}.
\newblock \DOIprefix\doi{https://doi.org/10.1016/j.compag.2022.107585}.
%Type = Article
\bibitem[{Qiao et~al.(2020)Qiao, Wang, Zhang and Pei}]{qiao2020detection}
\bibinfo{author}{Qiao, S.}, \bibinfo{author}{Wang, Q.}, \bibinfo{author}{Zhang,
  J.}, \bibinfo{author}{Pei, Z.}, \bibinfo{year}{2020}.
\newblock \bibinfo{title}{Detection and classification of early decay on
  blueberry based on improved deep residual 3d convolutional neural network in
  hyperspectral images}.
\newblock \bibinfo{journal}{Scientific Programming} \bibinfo{volume}{2020}.
\newblock \DOIprefix\doi{https://doi.org/10.1155/2020/8895875}.
%Type = Article
\bibitem[{Qin et~al.(2009)Qin, Burks, Ritenour and Bonn}]{qin2009detection}
\bibinfo{author}{Qin, J.}, \bibinfo{author}{Burks, T.F.},
  \bibinfo{author}{Ritenour, M.A.}, \bibinfo{author}{Bonn, W.G.},
  \bibinfo{year}{2009}.
\newblock \bibinfo{title}{Detection of citrus canker using hyperspectral
  reflectance imaging with spectral information divergence}.
\newblock \bibinfo{journal}{Journal of food engineering} \bibinfo{volume}{93},
  \bibinfo{pages}{183--191}.
\newblock \DOIprefix\doi{https://doi.org/10.1016/j.jfoodeng.2009.01.014}.
%Type = Book
\bibitem[{Rees(2013)}]{rees2013physical}
\bibinfo{author}{Rees, W.G.}, \bibinfo{year}{2013}.
\newblock \bibinfo{title}{Physical principles of remote sensing}.
\newblock \bibinfo{edition}{3} ed., \bibinfo{publisher}{Cambridge university
  press}.
\newblock \DOIprefix\doi{https://doi.org/10.1017/CBO9781139017411}.
%Type = Misc
\bibitem[{{Resonon Inc}(2023)}]{RI}
\bibinfo{author}{{Resonon Inc}}, \bibinfo{year}{2023}.
\newblock \bibinfo{title}{Spectronon software}.
\newblock \URLprefix \url{https://resonon.com/software}.
%Type = Article
\bibitem[{Rizzo et~al.(2021)Rizzo, Lichtveld, Mazet, Togami and
  Miller}]{rizzo2021plant}
\bibinfo{author}{Rizzo, D.M.}, \bibinfo{author}{Lichtveld, M.},
  \bibinfo{author}{Mazet, J.A.K.}, \bibinfo{author}{Togami, E.},
  \bibinfo{author}{Miller, S.A.}, \bibinfo{year}{2021}.
\newblock \bibinfo{title}{Plant health and its effects on food safety and
  security in a one health framework: four case studies}.
\newblock \bibinfo{journal}{One Health Outlook} \bibinfo{volume}{3}.
\newblock \DOIprefix\doi{10.1186/s42522-021-00038-7}.
%Type = Article
\bibitem[{Santos et~al.(2015)Santos, Guimaraes and Santos}]{dos2015efficient}
\bibinfo{author}{Santos, L.C.B.D.}, \bibinfo{author}{Guimaraes, S.J.F.},
  \bibinfo{author}{Santos, J.A.D.}, \bibinfo{year}{2015}.
\newblock \bibinfo{title}{Efficient unsupervised band selection through
  spectral rhythms}.
\newblock \bibinfo{journal}{Journal of Selected Topics in Signal Processing}
  \bibinfo{volume}{9}, \bibinfo{pages}{1016--1025}.
\newblock \DOIprefix\doi{https://doi.org/10.1109/JSTSP.2015.2405902}.
%Type = Article
\bibitem[{Sawant and Prabukumar(2020)}]{sawant2020survey}
\bibinfo{author}{Sawant, S.S.}, \bibinfo{author}{Prabukumar, M.},
  \bibinfo{year}{2020}.
\newblock \bibinfo{title}{A survey of band selection techniques for
  hyperspectral image classification}.
\newblock \bibinfo{journal}{Journal of Spectral Imaging} \bibinfo{volume}{9}.
\newblock \DOIprefix\doi{https://doi.org/10.1255/jsi.2020.a5}.
%Type = Article
\bibitem[{Selvaraju et~al.(2020)Selvaraju, Cogswell, Das, Vedantam, Parikh and
  Batra}]{selvaraju1610grad}
\bibinfo{author}{Selvaraju, R.R.}, \bibinfo{author}{Cogswell, M.},
  \bibinfo{author}{Das, A.}, \bibinfo{author}{Vedantam, R.},
  \bibinfo{author}{Parikh, D.}, \bibinfo{author}{Batra, D.},
  \bibinfo{year}{2020}.
\newblock \bibinfo{title}{Grad-cam: Visual explanations from deep networks via
  gradient-based localization}.
\newblock \bibinfo{journal}{International Journal of Computer Vision}
  \bibinfo{volume}{128}, \bibinfo{pages}{336--359}.
\newblock \DOIprefix\doi{https://doi.org/10.1007/s11263-019-01228-7}.
%Type = Article
\bibitem[{Shaikh et~al.(2021)Shaikh, Jaferzadeh, Th{\"o}rnberg and
  Casselgren}]{shaikh2021calibration}
\bibinfo{author}{Shaikh, M.S.}, \bibinfo{author}{Jaferzadeh, K.},
  \bibinfo{author}{Th{\"o}rnberg, B.}, \bibinfo{author}{Casselgren, J.},
  \bibinfo{year}{2021}.
\newblock \bibinfo{title}{Calibration of a hyper-spectral imaging system using
  a low-cost reference}.
\newblock \bibinfo{journal}{Sensors} \bibinfo{volume}{21},
  \bibinfo{pages}{3738}.
\newblock \DOIprefix\doi{https://doi.org/10.3390/s21113738}.
%Type = Techreport
\bibitem[{Sharma et~al.(2016)Sharma, Diba, Tuytelaars and
  Van~Gool}]{sharma2016hyperspectral}
\bibinfo{author}{Sharma, V.}, \bibinfo{author}{Diba, A.},
  \bibinfo{author}{Tuytelaars, T.}, \bibinfo{author}{Van~Gool, L.},
  \bibinfo{year}{2016}.
\newblock \bibinfo{title}{Hyperspectral CNN for image classification \& band
  selection, with application to face recognition}.
\newblock \bibinfo{type}{Technical Report}. KU Leuven.
%Type = Article
\bibitem[{Shi et~al.(2016)Shi, Gao, He, Li and Xu}]{shi2016hyperspectral}
\bibinfo{author}{Shi, A.}, \bibinfo{author}{Gao, H.}, \bibinfo{author}{He, Z.},
  \bibinfo{author}{Li, M.}, \bibinfo{author}{Xu, L.}, \bibinfo{year}{2016}.
\newblock \bibinfo{title}{A hyperspectral band selection based on game theory
  and differential evolution algorithm}.
\newblock \bibinfo{journal}{International Journal on Smart Sensing and
  Intelligent Systems} \bibinfo{volume}{9}, \bibinfo{pages}{1971--1990}.
\newblock \DOIprefix\doi{http://dx.doi.org/10.21307/ijssis-2017-948}.
%Type = Article
\bibitem[{Signoroni et~al.(2019)Signoroni, Savardi, Baronio and
  Benini}]{signoroni2019deep}
\bibinfo{author}{Signoroni, A.}, \bibinfo{author}{Savardi, M.},
  \bibinfo{author}{Baronio, A.}, \bibinfo{author}{Benini, S.},
  \bibinfo{year}{2019}.
\newblock \bibinfo{title}{Deep learning meets hyperspectral image analysis: A
  multidisciplinary review}.
\newblock \bibinfo{journal}{Journal of Imaging} \bibinfo{volume}{5},
  \bibinfo{pages}{52}.
\newblock \DOIprefix\doi{https://doi.org/10.3390/jimaging5050052}.
%Type = Article
\bibitem[{Simonyan et~al.(2013)Simonyan, Vedaldi and
  Zisserman}]{simonyan2013deep}
\bibinfo{author}{Simonyan, K.}, \bibinfo{author}{Vedaldi, A.},
  \bibinfo{author}{Zisserman, A.}, \bibinfo{year}{2013}.
\newblock \bibinfo{title}{Deep inside convolutional networks: Visualising image
  classification models and saliency maps}.
\newblock \bibinfo{journal}{arXiv} \bibinfo{volume}{abs/1312.6034}.
\newblock \DOIprefix\doi{https://doi.org/10.48550/arXiv.1312.6034}.
%Type = Article
\bibitem[{Su et~al.(2012)Su, Sheng, Du and Liu}]{su2012adaptive}
\bibinfo{author}{Su, H.}, \bibinfo{author}{Sheng, Y.}, \bibinfo{author}{Du,
  P.}, \bibinfo{author}{Liu, K.}, \bibinfo{year}{2012}.
\newblock \bibinfo{title}{Adaptive affinity propagation with spectral angle
  mapper for semi-supervised hyperspectral band selection}.
\newblock \bibinfo{journal}{Applied optics} \bibinfo{volume}{51},
  \bibinfo{pages}{2656--2663}.
\newblock \DOIprefix\doi{https://doi.org/10.1364/AO.51.002656}.
%Type = Article
\bibitem[{Su et~al.(2011)Su, Yang, Du and Sheng}]{su2011semisupervised}
\bibinfo{author}{Su, H.}, \bibinfo{author}{Yang, H.}, \bibinfo{author}{Du, Q.},
  \bibinfo{author}{Sheng, Y.}, \bibinfo{year}{2011}.
\newblock \bibinfo{title}{Semisupervised band clustering for dimensionality
  reduction of hyperspectral imagery}.
\newblock \bibinfo{journal}{IEEE Geoscience and Remote Sensing Letters}
  \bibinfo{volume}{8}, \bibinfo{pages}{1135--1139}.
\newblock \DOIprefix\doi{https://doi.org/10.1109/LGRS.2011.2158185}.
%Type = Book
\bibitem[{Sun(2010)}]{sun2010hyperspectral}
\bibinfo{author}{Sun, D.W.}, \bibinfo{year}{2010}.
\newblock \bibinfo{title}{Hyperspectral imaging for food quality analysis and
  control}.
\newblock \bibinfo{publisher}{Elsevier}.
\newblock \DOIprefix\doi{https://doi.org/10.1016/C2009-0-01853-4}.
%Type = Article
\bibitem[{Sun et~al.(2014a)Sun, Geng and Ji}]{sun2014efficient}
\bibinfo{author}{Sun, K.}, \bibinfo{author}{Geng, X.}, \bibinfo{author}{Ji,
  L.}, \bibinfo{year}{2014}a.
\newblock \bibinfo{title}{An efficient unsupervised band selection method based
  on an autocorrelation matrix for a hyperspectral image}.
\newblock \bibinfo{journal}{International Journal of Remote Sensing}
  \bibinfo{volume}{35}, \bibinfo{pages}{7458--7476}.
\newblock \DOIprefix\doi{https://doi.org/10.1080/01431161.2014.968686}.
%Type = Article
\bibitem[{Sun et~al.(2014b)Sun, Geng and Ji}]{sun2014new}
\bibinfo{author}{Sun, K.}, \bibinfo{author}{Geng, X.}, \bibinfo{author}{Ji,
  L.}, \bibinfo{year}{2014}b.
\newblock \bibinfo{title}{A new sparsity-based band selection method for target
  detection of hyperspectral image}.
\newblock \bibinfo{journal}{IEEE Geoscience and Remote Sensing Letters}
  \bibinfo{volume}{12}, \bibinfo{pages}{329--333}.
\newblock \DOIprefix\doi{https://doi.org/10.1109/LGRS.2014.2337957}.
%Type = Article
\bibitem[{Sun et~al.(2021)Sun, Song, Guo, Zhao and Wang}]{sun2021patch}
\bibinfo{author}{Sun, L.}, \bibinfo{author}{Song, X.}, \bibinfo{author}{Guo,
  H.}, \bibinfo{author}{Zhao, G.}, \bibinfo{author}{Wang, J.},
  \bibinfo{year}{2021}.
\newblock \bibinfo{title}{Patch-wise semantic segmentation for hyperspectral
  images via a cubic capsule network with emap features}.
\newblock \bibinfo{journal}{Remote Sensing} \bibinfo{volume}{13},
  \bibinfo{pages}{3497}.
\newblock \DOIprefix\doi{https://doi.org/10.3390/rs13173497}.
%Type = Article
\bibitem[{Sun and Du(2019)}]{sun2019hyperspectral}
\bibinfo{author}{Sun, W.}, \bibinfo{author}{Du, Q.}, \bibinfo{year}{2019}.
\newblock \bibinfo{title}{Hyperspectral band selection: A review}.
\newblock \bibinfo{journal}{IEEE Geoscience and Remote Sensing Magazine}
  \bibinfo{volume}{7}, \bibinfo{pages}{118--139}.
\newblock \DOIprefix\doi{https://doi.org/10.1109/MGRS.2019.2911100}.
%Type = Article
\bibitem[{Sun et~al.(2017)Sun, Tian, Xu, Zhang and Du}]{sun2017fast}
\bibinfo{author}{Sun, W.}, \bibinfo{author}{Tian, L.}, \bibinfo{author}{Xu,
  Y.}, \bibinfo{author}{Zhang, D.}, \bibinfo{author}{Du, Q.},
  \bibinfo{year}{2017}.
\newblock \bibinfo{title}{Fast and robust self-representation method for
  hyperspectral band selection}.
\newblock \bibinfo{journal}{IEEE Journal of Selected Topics in Applied Earth
  Observations and Remote Sensing} \bibinfo{volume}{10},
  \bibinfo{pages}{5087--5098}.
\newblock \DOIprefix\doi{https://doi.org/10.1109/JSTARS.2017.2737400}.
%Type = Inproceedings
\bibitem[{Tan and Lim(2019)}]{tan2019vanishing}
\bibinfo{author}{Tan, H.H.}, \bibinfo{author}{Lim, K.H.}, \bibinfo{year}{2019}.
\newblock \bibinfo{title}{Vanishing gradient mitigation with deep learning
  neural network optimization}, in: \bibinfo{booktitle}{Proceedings of
  International Conference on Smart Computing \& Communications (ICSCC)},
  \bibinfo{publisher}{IEEE}, \bibinfo{address}{Sarawak, Malaysia}. pp.
  \bibinfo{pages}{1--4}.
\newblock \DOIprefix\doi{https://doi.org/10.1109/ICSCC.2019.8843652}.
%Type = Article
\bibitem[{Tarabalka et~al.(2010)Tarabalka, Benediktsson, Chanussot and
  Tilton}]{tarabalka2010multiple}
\bibinfo{author}{Tarabalka, Y.}, \bibinfo{author}{Benediktsson, J.A.},
  \bibinfo{author}{Chanussot, J.}, \bibinfo{author}{Tilton, J.C.},
  \bibinfo{year}{2010}.
\newblock \bibinfo{title}{Multiple spectral--spatial classification approach
  for hyperspectral data}.
\newblock \bibinfo{journal}{IEEE Transactions on Geoscience and Remote Sensing}
  \bibinfo{volume}{48}, \bibinfo{pages}{4122--4132}.
\newblock \DOIprefix\doi{https://doi.org/10.1109/TGRS.2010.2062526}.
%Type = Inproceedings
\bibitem[{Tschannerl et~al.(2018)Tschannerl, Ren, Zabalza and
  Marshall}]{tschannerl2018segmented}
\bibinfo{author}{Tschannerl, J.}, \bibinfo{author}{Ren, J.},
  \bibinfo{author}{Zabalza, J.}, \bibinfo{author}{Marshall, S.},
  \bibinfo{year}{2018}.
\newblock \bibinfo{title}{Segmented autoencoders for unsupervised embedded
  hyperspectral band selection}, in: \bibinfo{booktitle}{Proceedings of
  European workshop on visual information processing (EUVIP)},
  \bibinfo{publisher}{IEEE}, \bibinfo{address}{Tampere, Finland}. pp.
  \bibinfo{pages}{1--6}.
\newblock \DOIprefix\doi{https://doi.org/10.1109/EUVIP.2018.8611643}.
%Type = Inproceedings
\bibitem[{Uzair and Jamil(2020)}]{uzair2020effects}
\bibinfo{author}{Uzair, M.}, \bibinfo{author}{Jamil, N.}, \bibinfo{year}{2020}.
\newblock \bibinfo{title}{Effects of hidden layers on the efficiency of neural
  networks}, in: \bibinfo{booktitle}{IEEE 23rd international multitopic
  conference (INMIC)}, \bibinfo{organization}{IEEE}. pp. \bibinfo{pages}{1--6}.
\newblock \DOIprefix\doi{https://doi.org/10.1109/INMIC50486.2020.9318195}.
%Type = Article
\bibitem[{Vaiphasa(2006)}]{vaiphasa2006consideration}
\bibinfo{author}{Vaiphasa, C.}, \bibinfo{year}{2006}.
\newblock \bibinfo{title}{Consideration of smoothing techniques for
  hyperspectral remote sensing}.
\newblock \bibinfo{journal}{ISPRS journal of photogrammetry and remote sensing}
  \bibinfo{volume}{60}, \bibinfo{pages}{91--99}.
\newblock \DOIprefix\doi{https://doi.org/10.1016/j.isprsjprs.2005.11.002}.
%Type = Article
\bibitem[{Vidal and Amigo(2012)}]{vidal2012pre}
\bibinfo{author}{Vidal, M.}, \bibinfo{author}{Amigo, J.M.},
  \bibinfo{year}{2012}.
\newblock \bibinfo{title}{Pre-processing of hyperspectral images. essential
  steps before image analysis}.
\newblock \bibinfo{journal}{Chemometrics and Intelligent Laboratory Systems}
  \bibinfo{volume}{117}, \bibinfo{pages}{138--148}.
\newblock \DOIprefix\doi{https://doi.org/10.1016/j.chemolab.2012.05.009}.
%Type = Article
\bibitem[{Wan et~al.(2023)Wan, Lu, Wang, Shen, Xu and Lang}]{wan2023yolo}
\bibinfo{author}{Wan, D.}, \bibinfo{author}{Lu, R.}, \bibinfo{author}{Wang,
  S.}, \bibinfo{author}{Shen, S.}, \bibinfo{author}{Xu, T.},
  \bibinfo{author}{Lang, X.}, \bibinfo{year}{2023}.
\newblock \bibinfo{title}{Yolo-hr: Improved yolov5 for object detection in
  high-resolution optical remote sensing images}.
\newblock \bibinfo{journal}{Remote Sensing} \bibinfo{volume}{15},
  \bibinfo{pages}{614}.
\newblock \DOIprefix\doi{https://doi.org/10.3390/rs15030614}.
%Type = Article
\bibitem[{Wang et~al.(2021)Wang, Liu, Liu, Zhu, Hou, Liu and
  Li}]{wang2021review}
\bibinfo{author}{Wang, C.}, \bibinfo{author}{Liu, B.}, \bibinfo{author}{Liu,
  L.}, \bibinfo{author}{Zhu, Y.}, \bibinfo{author}{Hou, J.},
  \bibinfo{author}{Liu, P.}, \bibinfo{author}{Li, X.}, \bibinfo{year}{2021}.
\newblock \bibinfo{title}{A review of deep learning used in the hyperspectral
  image analysis for agriculture}.
\newblock \bibinfo{journal}{Artificial Intelligence Review}
  \bibinfo{volume}{54}, \bibinfo{pages}{5205--5253}.
\newblock \DOIprefix\doi{https://doi.org/10.1007/s10462-021-10018-y}.
%Type = Article
\bibitem[{Wang et~al.(2007)Wang, Jia and Zhang}]{wang2007novel}
\bibinfo{author}{Wang, L.}, \bibinfo{author}{Jia, X.}, \bibinfo{author}{Zhang,
  Y.}, \bibinfo{year}{2007}.
\newblock \bibinfo{title}{A novel geometry-based feature-selection technique
  for hyperspectral imagery}.
\newblock \bibinfo{journal}{IEEE Geoscience and Remote Sensing Letters}
  \bibinfo{volume}{4}, \bibinfo{pages}{171--175}.
\newblock \DOIprefix\doi{https://doi.org/10.1109/LGRS.2006.887142}.
%Type = Article
\bibitem[{Yang et~al.(2017)Yang, Tan, Bruzzone, Lu and
  Guan}]{yang2017discriminative}
\bibinfo{author}{Yang, C.}, \bibinfo{author}{Tan, Y.},
  \bibinfo{author}{Bruzzone, L.}, \bibinfo{author}{Lu, L.},
  \bibinfo{author}{Guan, R.}, \bibinfo{year}{2017}.
\newblock \bibinfo{title}{Discriminative feature metric learning in the
  affinity propagation model for band selection in hyperspectral images}.
\newblock \bibinfo{journal}{Remote Sensing} \bibinfo{volume}{9},
  \bibinfo{pages}{782}.
\newblock \DOIprefix\doi{https://doi.org/10.3390/rs9080782}.
%Type = Incollection
\bibitem[{Yang et~al.(2009)Yang, Jin and Yang}]{Yang2009}
\bibinfo{author}{Yang, J.}, \bibinfo{author}{Jin, Z.}, \bibinfo{author}{Yang,
  J.}, \bibinfo{year}{2009}.
\newblock \bibinfo{title}{Non-linear techniques for dimension reduction}, in:
  \bibinfo{booktitle}{Encyclopedia of Biometrics}.
  \bibinfo{publisher}{Springer}, pp. \bibinfo{pages}{1003--1007}.
\newblock \DOIprefix\doi{https://doi.org/10.1007/978-0-387-73003-5_294}.
%Type = Inproceedings
\bibitem[{Yin et~al.(2010)Yin, Wang and Zhao}]{yin2010optimal}
\bibinfo{author}{Yin, J.}, \bibinfo{author}{Wang, Y.}, \bibinfo{author}{Zhao,
  Z.}, \bibinfo{year}{2010}.
\newblock \bibinfo{title}{Optimal band selection for hyperspectral image
  classification based on inter-class separability}, in:
  \bibinfo{booktitle}{Proceedings of Symposium on Photonics and
  Optoelectronics}, \bibinfo{publisher}{IEEE}, \bibinfo{address}{Chengdu,
  China}. pp. \bibinfo{pages}{1--4}.
\newblock \DOIprefix\doi{https://doi.org/10.1109/SOPO.2010.5504325}.
%Type = Article
\bibitem[{Yin et~al.(2020)Yin, Wang and Yang}]{yin2020novel}
\bibinfo{author}{Yin, S.}, \bibinfo{author}{Wang, Y.}, \bibinfo{author}{Yang,
  Y.H.}, \bibinfo{year}{2020}.
\newblock \bibinfo{title}{A novel image-dehazing network with a parallel
  attention block}.
\newblock \bibinfo{journal}{Pattern Recognition} \bibinfo{volume}{102},
  \bibinfo{pages}{107255}.
\newblock \DOIprefix\doi{https://doi.org/10.1016/j.patcog.2020.107255}.
%Type = Inproceedings
\bibitem[{Zhai et~al.(2016)Zhai, Zhang, Zhang and Li}]{zhai2016squaring}
\bibinfo{author}{Zhai, H.}, \bibinfo{author}{Zhang, H.},
  \bibinfo{author}{Zhang, L.}, \bibinfo{author}{Li, P.}, \bibinfo{year}{2016}.
\newblock \bibinfo{title}{Squaring weighted low-rank subspace clustering for
  hyperspectral image band selection}, in: \bibinfo{booktitle}{Proceedings of
  International Geoscience and Remote Sensing Symposium (IGARSS)},
  \bibinfo{publisher}{IEEE}, \bibinfo{address}{Beijing, China}. pp.
  \bibinfo{pages}{2434--2437}.
\newblock \DOIprefix\doi{https://doi.org/10.1109/IGARSS.2016.7729628}.
%Type = Article
\bibitem[{Zhan et~al.(2017)Zhan, Hu, Xing and Yu}]{zhan2017hyperspectral}
\bibinfo{author}{Zhan, Y.}, \bibinfo{author}{Hu, D.}, \bibinfo{author}{Xing,
  H.}, \bibinfo{author}{Yu, X.}, \bibinfo{year}{2017}.
\newblock \bibinfo{title}{Hyperspectral band selection based on deep
  convolutional neural network and distance density}.
\newblock \bibinfo{journal}{IEEE Geoscience and Remote Sensing Letters}
  \bibinfo{volume}{14}, \bibinfo{pages}{2365--2369}.
\newblock \DOIprefix\doi{https://doi.org/10.1109/LGRS.2017.2765339}.
%Type = Inproceedings
\bibitem[{Zhou et~al.(2016)Zhou, Khosla, Lapedriza, Oliva and
  Torralba}]{zhou2016learning}
\bibinfo{author}{Zhou, B.}, \bibinfo{author}{Khosla, A.},
  \bibinfo{author}{Lapedriza, A.}, \bibinfo{author}{Oliva, A.},
  \bibinfo{author}{Torralba, A.}, \bibinfo{year}{2016}.
\newblock \bibinfo{title}{Learning deep features for discriminative
  localization}, in: \bibinfo{booktitle}{Proceedings of Conference on Computer
  Vision and Pattern Recognition ({CVPR})}, \bibinfo{publisher}{IEEE},
  \bibinfo{address}{Las Vegas, NV, USA}. pp. \bibinfo{pages}{2921--2929}.
\newblock \DOIprefix\doi{https://doi.org/10.1109/CVPR.2016.319}.

\end{thebibliography}
		
	\end{document}